\documentclass[12pt,a4paper]{article}
\usepackage{graphicx}
\usepackage{geometry}
\usepackage{booktabs}
\usepackage{threeparttable} 
\usepackage{cite}
\usepackage{amsmath,amssymb}
\usepackage{hyperref}
\usepackage{multirow}
\usepackage{array}
\usepackage{xcolor}
\usepackage{colortbl}
\usepackage{longtable}
\usepackage{pdflscape}
\usepackage{caption}
\usepackage{subcaption}
\usepackage{float}
\usepackage{xurl}
\usepackage{enumitem}
\usepackage{microtype}

\usepackage{chngcntr}
\counterwithin{figure}{section}
\counterwithin{table}{section}

\usepackage{iftex}
\ifXeTeX
  \usepackage{fontspec}
  \newfontfamily\banglafont[Script=Bengali]{Kalpurush}
\else
  \providecommand{\banglafont}{}
\fi
\ifdefined\DeclareUnicodeCharacter
  \DeclareUnicodeCharacter{0964}{\textbar}
\fi

\begin{document}

\title{UNIFIED MULTI-DIALECTAL NEURAL MACHINE TRANSLATION FOR BANGLA USING THE DWADASH BENCHMARK CORPUS}
\author{Tanbir Ahmed \and Rakib Ullah \and Md. Ruhul Islam \and Nayan Kumar Nath \\ Sylhet Engineering College}
\date{}
\maketitle

\begin{abstract}
While Neural Machine Translation (NMT) and Large Language Models (LLMs) excel at cross-lingual tasks, they frequently fail to capture intra-lingual morphological variation, marginalizing millions of dialectal speakers. In Bangla, existing translation frameworks exacerbate this issue by relying on Standard Colloquial Bangla (SCB) as an intermediate pivot, which compounds translation errors and erodes cross-dialectal nuance. To bridge this dialectal divide, this paper introduces a unified, multi-directional NMT system capable of direct translation across SCB and eleven regional variants. We first present a comprehensive literature review that systematically synthesizes prior dialectal NLP resources to map existing technological gaps. As a foundational contribution, we construct and release the largest multi-dialect parallel corpus for Bangla to date—comprising 14,562 aligned rows and yielding 51,541 non-null sentence pairs through the integration of seven prior datasets and targeted, native-speaker-verified manual augmentation. Leveraging this large-scale corpus, we systematically benchmark state-of-the-art sequence-to-sequence architectures using parameter-efficient Weight-Decomposed Low-Rank Adaptation (DoRA). Empirical results demonstrate that deep monolingual pre-training is more effective than massive multilingual capacity for this task: the compact BanglaT5 model outperforms NLLB-200 and mBART-50 by up to +13.96 BLEU, achieving state-of-the-art scores of 29.26 BLEU, 57.26 chrF++, and 49.68 METEOR. Furthermore, our dataset scaling study reveals that translation quality yields diminishing returns beyond 3,000 parallel pairs, and that linguistic proximity to Standard Bangla ultimately outweighs raw data volume in determining final translation efficacy. Finally, we deploy the optimized model as an INT8-quantized web application, establishing a scalable, open-source framework for inclusive language technology and equitable digital access.

\vspace{1cm}

\noindent \textbf{Keywords:} Neural Machine Translation, Poly-Dialectal Translation, Bangla Regional Dialects, Large Language Models, Sequence-to-Sequence Architectures, Low-Rank Adaptation (DoRA).
\end{abstract}

\section{Introduction}
\subsection{Background and Motivation}
Bangla (Bengali) belongs to the Indo-Aryan branch of the Indo-European language family and ranks among the most widely spoken languages in the world, serving as the mother tongue of more than 240 million people across Bangladesh and the neighbouring Indian states of West Bengal, Tripura, and Assam. Although the language enjoys a rich literary tradition and considerable socio-political recognition, it is far from monolithic. Numerous regional dialects---including Sylheti, Chittagonian, Barisali, Noakhali, Rangpuri, Rajshahi, and Mymensingh---diverge from Standard Colloquial Bangla (SCB) across phonological, morphological, syntactic, and lexical dimensions. For certain peripheral variants, such as Chittagonian and Sylheti, these divergences are so pronounced that mutual intelligibility with Standard Bangla is severely limited \cite{faria2025vashantor}.

This dialectal diversity, while culturally significant, creates tangible communication barriers. Speakers of regional dialects often struggle in formal settings---legal proceedings, educational institutions, and administrative interactions---that are conducted exclusively in SCB. The problem extends into the digital domain: contemporary Natural Language Processing (NLP) tools, machine translation services, and conversational interfaces are overwhelmingly built for Standard Bangla, effectively excluding millions of dialectal speakers from full participation in the digital economy. The resulting gap in language technology underscores a pressing need for inclusive NLP systems that can serve dialectal communities without demanding linguistic assimilation.

\subsection{Problem Statement}
While Neural Machine Translation (NMT) architectures and Large Language Models (LLMs) have achieved remarkable performance on cross-lingual tasks between high-resource languages, their effectiveness diminishes sharply when the translation boundary lies \textit{within} a single language---between its standard register and regional dialects. Nearly all existing Bangla NLP benchmarks and pre-trained models assume a homogeneous language distribution, thereby disregarding the rich cross-dialectal variation that characterises actual language use.

Three interrelated technical obstacles underpin this shortcoming. First, parallel corpora spanning multiple dialects remain scarce, depriving fine-tuning pipelines of the supervised signal needed for robust generalisation. Second, the absence of a standardised orthographic convention for regional dialects amplifies tokenisation errors: subword segmentation algorithms such as BPE \cite{sennrich2015neural} and WordPiece \cite{wu2016google}, calibrated on SCB text, tend to split dialectal tokens into fragments that lose their original morphological coherence. Third, the few available dialectal datasets exhibit severe class imbalance and cater primarily to classification tasks (e.g., dialect identification, sentiment analysis) rather than the sequence-to-sequence alignment required for generative translation.

\begin{figure}[htbp]
  \centering
  \includegraphics[width=\textwidth]{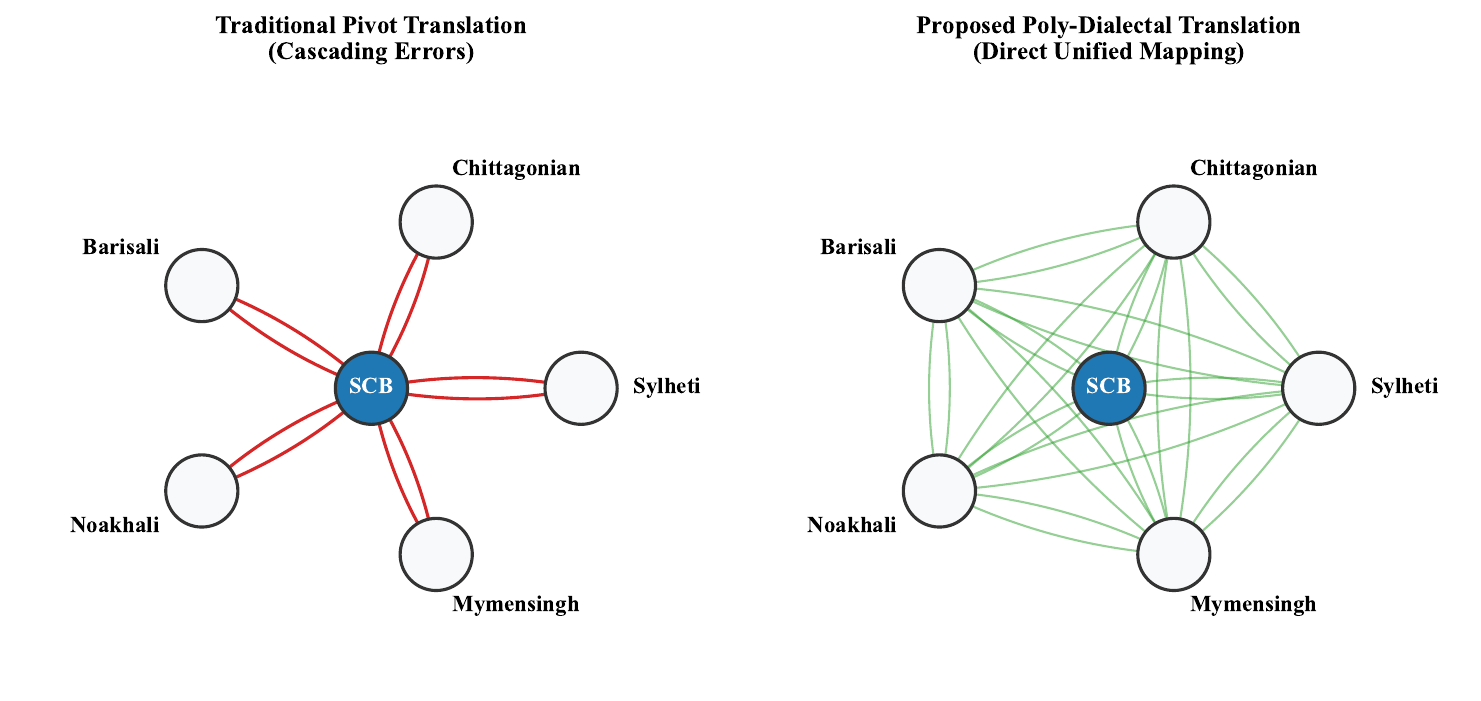}
  \caption{Structural comparison between the traditional pivot translation approach (routing through Standard Bangla) and the proposed direct poly-dialectal mapping topology.}
  \label{fig:topology}
\end{figure}

Formally, the poly-dialectal translation problem is stated as follows. Let $\mathcal{D} = \{d_1, \allowbreak d_2, \allowbreak \ldots, \allowbreak d_k\}$ denote a set of $k$ regional dialect variants (including SCB). Given a source sentence $\mathbf{x} = (x_1, x_2, \ldots, x_m)$ in dialect $d_s \in \mathcal{D}$ and a target dialect $d_t \in \mathcal{D}$ where $d_s \neq d_t$, the goal is to learn a unified mapping $\mathcal{T}_\theta: (\mathbf{x}, d_s, d_t) \rightarrow \mathbf{y}$ that produces a target sentence $\mathbf{y} = (y_1, y_2, \ldots, y_n)$ preserving the semantic content of $\mathbf{x}$ while adopting the morphological and lexical conventions of $d_t$. Meeting this goal demands a shift from isolated, uni-dialectal modelling to a unified, poly-dialectal approach (see Figure~\ref{fig:topology}).

\subsection{Proposed System and Contributions}
\begin{figure}[htbp]
  \centering
  \includegraphics[width=.95\textwidth]{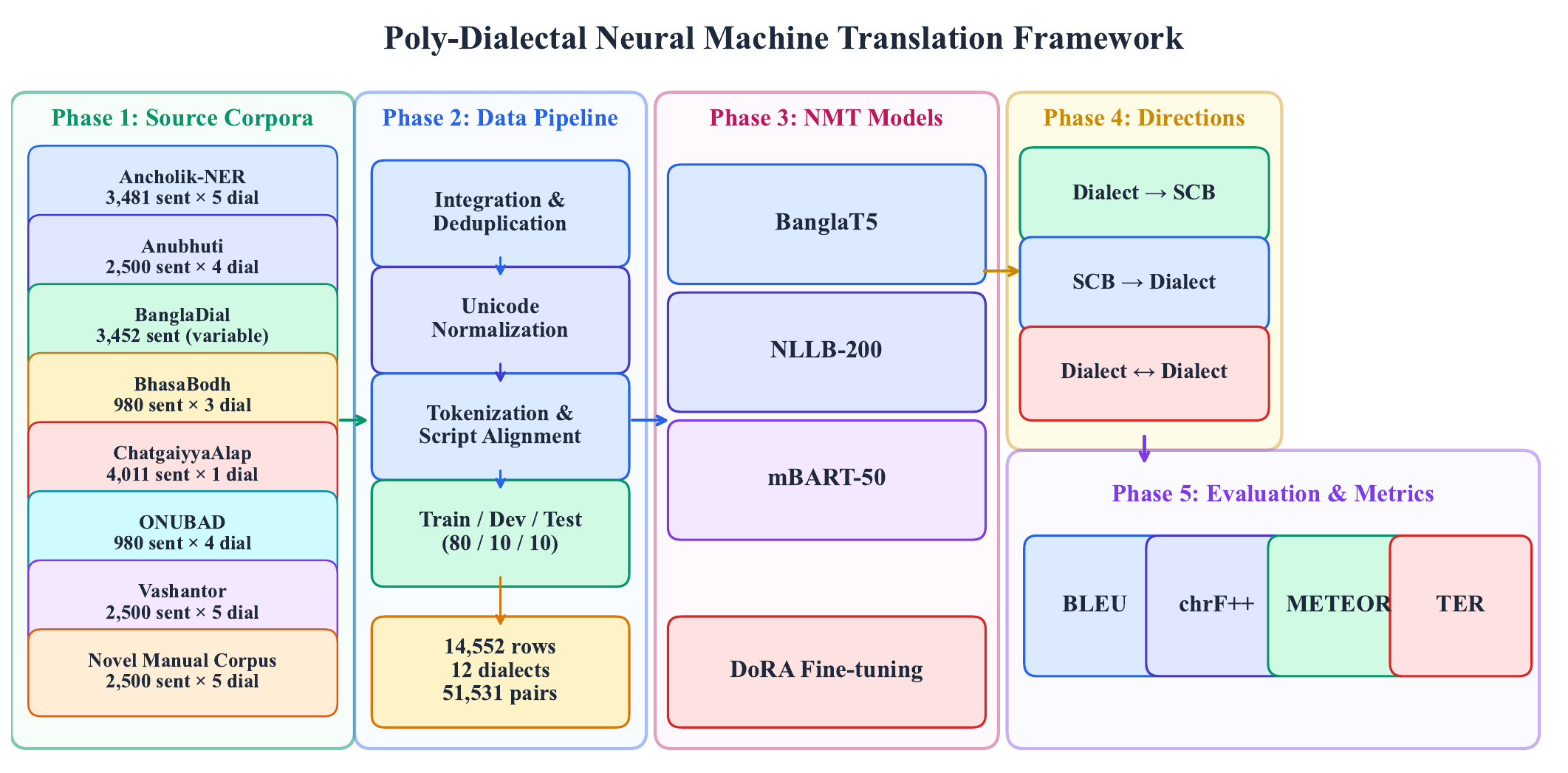}
  \caption{End-to-end architectural workflow of the Poly-Dialectal Neural Machine Translation Framework, depicting five phases: source corpora integration from seven existing datasets and one novel manual corpus, the data processing pipeline, NMT model fine-tuning with DoRA, multi-directional translation paradigm, and evaluation with four standard MT metrics.}
  \label{fig:overview}
\end{figure}

To overcome the limitations outlined above, this research introduces a unified \textbf{Poly-Dialectal Neural Machine Translation System} designed to accommodate the morphological and syntactic particularities of Bangla regional dialects. As illustrated in Figure~\ref{fig:overview}, the proposed architecture supports three translation paradigms within a single model: Dialect $\rightarrow$ Standard Bangla, Standard Bangla $\rightarrow$ Dialect, and direct Dialect $\rightarrow$ Dialect translation. By enabling direct inter-dialectal conversion, the system avoids the cascading error accumulation inherent in pivot-based approaches that route all translations through SCB.

The primary contributions of this paper are summarized as follows:
\begin{enumerate}[leftmargin=*]
    \item \textbf{Largest Multi-Dialectal Parallel Corpus:} We construct the most extensive multi-dialect machine translation corpus for Bangla to date, comprising 14,562 aligned rows across 12 regional dialect columns. The corpus integrates seven prior datasets---Ancholik-NER, Anubhuti, BanglaDial, BhasaBodh, ChatgaiyyaAlap, ONUBAD, and Vashantor---and incorporates 2,510 manually created, bidirectional, expert-verified parallel sentence pairs for five previously unaddressed dialects (Rangpur, Tangail, Kishoreganj, Narail, Narsingdi), yielding 51,541 non-null parallel pairs across 12 dialects. To support reproducible research and open access, our compiled dataset is publicly available at Mendeley Data (\url{https://data.mendeley.com/datasets/v9cf66fk2t}).
    \item \textbf{Systematic Benchmarking of NMT Architectures for Poly-Dialectal Translation:} To the best of our knowledge, this work presents the first systematic empirical evaluation of prominent sequence-to-sequence architectures---specifically BanglaT5 \cite{bhattacharjee-etal-2023-banglanlg}, NLLB-200 \cite{nllb2022}, and mBART-50 \cite{tang2020multilingual}---on the task of multi-directional, poly-dialectal Bangla translation under parameter-efficient fine-tuning.
    \item \textbf{Cross-Dialectal Transfer Analysis and Dataset Scaling Study:} We conduct a rigorous dataset scaling study (progressively scaling from 500 to 4,499 parallel instances) and quantitative characterization of cross-dialectal transfer mechanisms, revealing how morphological proximity and training data volume jointly influence translation efficacy across diverse evaluation metrics.
    \item \textbf{Real-World Web Deployment:} We deploy the best-performing NMT model as a publicly accessible, INT8-quantized web application, enabling real-time poly-dialectal translation across 12 dialects and bridging the digital language divide for marginalized dialect speakers.
    \item \textbf{Comprehensive Multi-Metric Evaluation Framework:} Unlike prior studies that predominantly evaluate translation quality using a single metric (typically BLEU) or at most two (BLEU and chrF++), we implement a more comprehensive evaluation protocol. By incorporating METEOR (semantic and morphological alignment) and Translation Error Rate (TER, post-editing effort) alongside standard metrics, we provide a multi-dimensional assessment of syntactic accuracy, semantic preservation, and practical post-editing utility.
\end{enumerate}

\subsection{Paper Organization}
The remainder of this paper is organized as follows. Section~2 critically reviews related literature on Bangla regional NLP and dialectal machine translation. Section~3 formally states the research hypothesis and objectives. Section~4 details the methodology, encompassing dataset construction, model architectures, and fine-tuning configurations. Section~5 presents the experimental results, including cross-dialectal analysis, error taxonomy, dataset scaling findings, and comparison with prior studies. Section~6 describes the deployment architecture and the publicly available web application. Section~7 discusses limitations and proposes future research directions. Section~8 concludes the paper.

\section{Literature Review}

\subsection{Overview of Regional Bangla NLP}
Historically, computational linguistic resources for Bangla have centred on the standard register, leaving regional dialects severely underrepresented. Over the past several years, however, a growing body of work has begun to address this imbalance, with early efforts concentrating on resource creation for text classification and sentiment analysis.

Sultana et al.~\cite{sultana2024bdregion} compiled the \textit{BdRegionText} corpus for regional text classification and showed that classifiers trained on standard Bangla exhibit marked performance degradation when confronted with dialectal vocabulary; their best-performing pipeline (Random Forest with TF-IDF features) reached only 79.15\% accuracy. Extending this line of inquiry, Mahi et al.~\cite{mahi2025bangladial} released \textit{BanglaDial}, a consolidated dataset spanning 12 dialects with 60,729 sentences. Their statistical characterisation of the corpus quantified the severe class imbalance that pervades available dialectal data, reporting an Imbalance Ratio of $IR = 9.82$ and a Shannon Entropy of $H = 3.416$ bits.

Beyond identification and classification, several studies have explored specialised downstream tasks in which dialectal morphology poses additional challenges. Kundu et al.~\cite{kundu2026anubhuti} introduced \textit{ANUBHUTI} for sentiment analysis across four regional variants, achieving high inter-rater agreement (Cohen's $\kappa = 0.76$--$0.84$) and documenting how emotional markers vary across dialect boundaries. Fayaz et al.~\cite{fayaz2025bidwesh} demonstrated through their \textit{BIDWESH} hate-speech detection corpus that toxicity classifiers trained on SCB fail to generalise to dialectal slang---a finding with direct implications for large-scale content moderation. Paul et al.~\cite{paul2025ancholik} contributed \textit{ANCHOLIK-NER}, the first set of dialectal named entity annotations for Sylheti, Chittagonian, and Barisali text. A common thread across all three studies is that they target classification objectives and do not furnish the generative capabilities needed for cross-dialectal translation.

\subsection{Dialectal Machine Translation and LLM Benchmarks}
While the classification-oriented resources reviewed above have broadened the understanding of dialectal Bangla text, generative machine translation between dialects has emerged as a distinct research focus only recently, catalysed by the availability of massively multilingual pre-trained models.

Among the earliest translation corpora, Chowdhury et al.~\cite{chowdhury2025chatgaiyya} constructed \textit{ChatgaiyyaAlap}, a parallel dataset of 4,012 sentences for Chittagonian-to-Standard Bangla conversion. Although valuable for the Chittagonian community, the dataset is inherently limited to a single dialect pair and cannot support multi-directional or cross-dialectal translation. Sultana et al.~\cite{sultana2025onubad} presented \textit{ONUBAD} with 7,950 parallel rows spanning three dialects and used it to establish initial NMT baselines with BanglaT5 and SeamlessM4T. Faria et al.~\cite{faria2025vashantor} introduced \textit{Vashantor}, a multilingual benchmark covering five dialects; their custom DialectBanglaT5 model attained 71.93 BLEU on Mymensingh translation. Nonetheless, \textit{Vashantor} was capped at 2,500 sentences per dialect---a volume that limits the robust generalisation required for diverse linguistic phenomena.

More recently, the research community has explored the capabilities of Large Language Models (LLMs) for dialect translation. Mahjabin et al.~\cite{mahjabin2025banglachq} constructed \textit{BanglaCHQ-Prantik} to benchmark LLMs in the medical domain, where Gemini 2.5 Flash achieved 23.67 BLEU on Sylheti translation. Jawad et al.~\cite{jawad2025dialtsa} presented \textit{DIALTSA-BN}, revealing that transliteration preprocessing substantially improves LLM translation performance (BLEU from $<$0.07 to 0.330 with Claude 3.7). Bhuiyan et al.~\cite{bhuiyan2025bhasabodh} explored \textit{BhasaBodh}, investigating script normalization and romanization for dialect translation, with mBART-50 achieving 87.44 BLEU on Romanized-to-Standard conversion. Despite these advances, a consistent finding across all LLM-based studies is that zero-shot and few-shot LLM approaches remain inferior to dedicated fine-tuned NMT models for specialized low-resource dialectal translation, owing to the absence of dialect-specific representations in pre-training data.

\subsection{Critical Synthesis and Research Gap}
The studies reviewed above have collectively laid a foundation for Bangla dialectal NLP. A critical synthesis, however, reveals four structural and methodological shortcomings that impede the development of production-ready translation systems:

\begin{enumerate}[leftmargin=*]
    \item \textbf{Single-Dialect Isolation:} Corpora such as \textit{ChatgaiyyaAlap} address a single dialect pair (Chittagonian$\leftrightarrow$Standard), and models trained on such isolated datasets lack the capacity to generalize across other regional dialects, necessitating separate models for each region.
    \item \textbf{Corpus Volume Limitations and Absence of Scaling Studies:} Multi-dialect datasets such as \textit{Vashantor} provide only 2,500 samples per dialect---a volume demonstrably insufficient for deep fine-tuning of architectures such as mT5 or mBART-50. Moreover, prior literature entirely lacks systematic scaling studies that empirically determine optimal dataset size thresholds for cross-dialectal transfer learning.
    \item \textbf{The Pivot Translation Problem:} No existing framework supports direct translation between different regional dialects (e.g., Sylheti $\rightarrow$ Chittagonian). Current systems require routing through Standard Bangla as an intermediary, which compounds translation errors and degrades semantic accuracy at each step.
    \item \textbf{Absence of Deployed Applications:} Models developed in prior studies remain confined to offline evaluation environments. The community lacks open, deployed web applications capable of providing real-time translation services to dialect speakers.
\end{enumerate}

\begin{figure}[htbp]
  \centering
  \includegraphics[width=\textwidth]{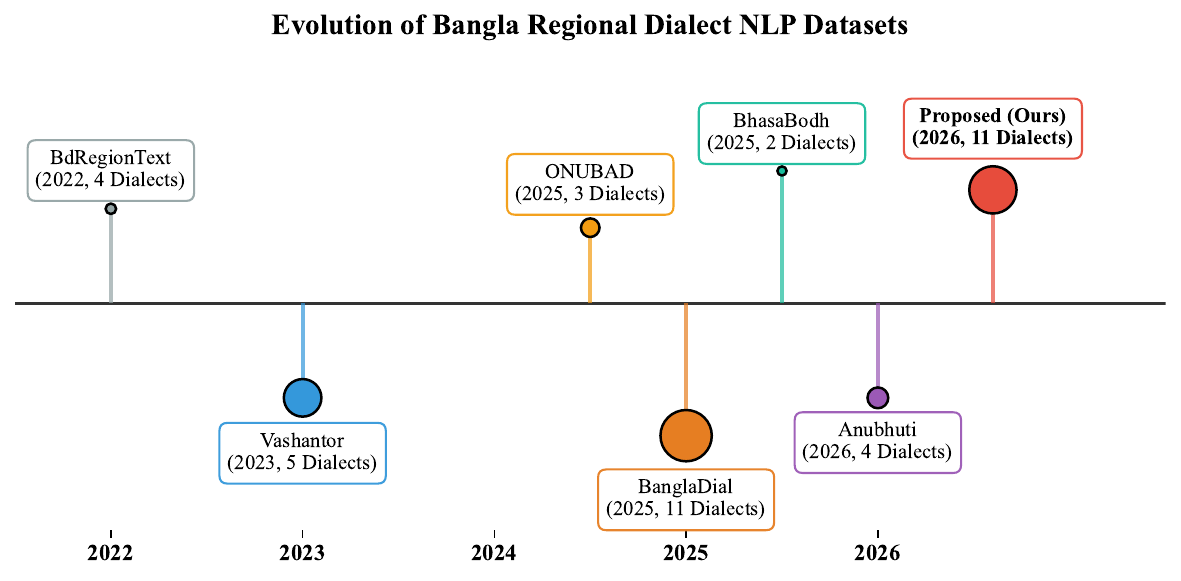}
  \caption{Evolution of Bangla regional dialect datasets and resources over time, highlighting the scale and dialectal coverage of the proposed corpus.}
  \label{fig:timeline}
\end{figure}

The present work addresses all four limitations by introducing a comprehensive poly-dialectal parallel corpus substantially larger than previous efforts (as visualized in Figure~\ref{fig:timeline}), enabling direct multi-directional translation within a single unified neural architecture, conducting the first systematic dataset scaling study, and deploying the resulting system as a publicly accessible web application.

\subsection{Summary of Dialectal Resources}
Table \ref{tab:literature_summary} summarizes the major dialectal datasets and resources reviewed in this section, detailing their coverage, architecture, evaluation metrics, and state-of-the-art performance.

{\small
{\small
\setlength{\tabcolsep}{4pt}
\begin{longtable}{
    >{\raggedright\arraybackslash}p{2.0cm}
    >{\raggedright\arraybackslash}p{2.3cm}
    >{\raggedright\arraybackslash}p{2.4cm}
    >{\raggedright\arraybackslash}p{3.2cm}
    >{\raggedright\arraybackslash}p{4.2cm}
}
\caption{Comprehensive Summary of Bangla Regional Dialect NLP Resources and Datasets}
\label{tab:literature_summary} \\
\toprule
\textbf{Dataset \& Year} & \textbf{Primary Task \& Domain} & \textbf{Dialects \& Dataset Scale} & \textbf{Methodology \& Quality Control} & \textbf{Key Metrics \& SOTA Findings} \\
\midrule
\endfirsthead

\multicolumn{5}{c}{{\bfseries \tablename\ \thetable{} -- continued from previous page}} \\
\toprule
\textbf{Dataset \& Year} & \textbf{Primary Task \& Domain} & \textbf{Dialects \& Dataset Scale} & \textbf{Methodology \& Quality Control} & \textbf{Key Metrics \& SOTA Findings} \\
\midrule
\endhead

\midrule
\multicolumn{5}{r}{{Continued on next page}} \\
\bottomrule
\endfoot

\bottomrule
\endlastfoot

Ancholik-Ner (2025) & Dialectal Named Entity Recognition (NER); 10 entity classes. & Sylhet, Chittagong, Barishal. 10,443 sentences (3,481/dialect). & BIO tagging. Sourced from Vashantor (71.8\%), ONUBAD, \& news. 3-annotator consensus pipeline. & First dialectal NER baseline. Assamese NER baseline reached 80.69\% F1 (MuRIL); medical BanglishBERT got 56.13\% F1. \\
\midrule
Anubhuti (2026) & Thematic Sentiment \& Multilabel Emotion Analysis in dialectal text. & Mymensingh, Noakhali, Sylhet, Chittagong. 10,000 sentences (2,500/region). & Sourced from MONOVAB. 8 native translators (2/region) with 4-stage QA (sentiment, NaN parsing, anomaly, spelling). & High inter-rater agreement (Cohen's $\kappa = 0.76$--$0.84$). Dominant emotions: Contempt (921), Disgust (871), Anger (732). \\
\midrule
Bidwesh (2025) & Multi-Dialectal Hate Speech \& Toxicity Detection. & Barishal, Noakhali, Chittagong. 9,183 instances (3,061 parallel/dialect). & Sourced from BD-SHS; translated by 6 native speakers; 5-stage validation. Target/Type taxonomy. & Balanced binary split (49.2\% Hate / 50.8\% Non-Hate). Individual targets highest (34.4\%); Slander dominant (53.7\%). \\
\midrule
Bangla Dialect Bibliography & Systematic Bibliography \& Meta-Review of Bangla Dialect NLP. & Comprehensive overview across major regional variants in BD. & Systematic synthesis tracking 11 publications, annotation strategies, \& baselines. & Mapped 11 unique corpora; highlights Sylheti \& Chittagonian as most studied; Barishal, Mymensingh, Noakhali emerging. \\
\midrule
BanglaDial (2025) & Dialect Identification (DID) \& Class Imbalance Modeling. & 12 Dialects. & Aggregated from 4 corpora (Vashantor, ONUBAD, etc.). Deduplicated (4.1\% removed) \& normalized. & Long-tail distribution (Top-3: 41.3\% of records). Imbalance Ratio $IR = 9.82$; Shannon Entropy $H = 3.416$ bits. \\
\midrule
BdRegion\newline Text (2022) & Bangla Regional Text Classification on social media content. & Chittagong, Noakhali, Barishal, Rangpur. 2,573 sample texts. & Scraped from FB/YouTube/IG with human validation. Extracted TF-IDF, CountVectorizer, \& BoW. & Random Forest + TF-IDF achieved peak 79.15\% accuracy (10-fold CV: 79.47\%, weighted F1: 0.78). RF + BoW got 75.92\%. \\
\midrule
BhasaBodh (2025) & Script Normalization \& Multilingual Dialect Translation (Native/Romanized). & Chittagong, Sylhet (aligned with Standard Bangla \& English). 1,960 sentences. & Sourced from ONUBAD; augmented via Gemini 2.5 Pro for Romanizations. Fine-tuned NLLB-200 \& mBART-50. & mBART-50 achieved peak 87.44 BLEU on Romanized $\rightarrow$ Standard Bangla and 74.36 BLEU on Chittagong $\leftrightarrow$ Sylhet. \\
\midrule
Chatgaiyya\newline Alap (2025) & Dialect Conversion (Chittagonian $\rightarrow$ Standard Bangla). & Chittagonian (aligned with Standard Bangla). 4,012 sentences + 1,500-word dictionary. & Sourced from social media/dramas; validated by 5 native Chittagonian graduates. & Created clean translation mapping resolving severe grammatical divergences \& phonetic shifts in negative sentences. \\
\midrule
BanglaCHQ-Prantik (2025) & Medical-Domain Machine Translation (Standard Bangla $\rightarrow$ Dialects). & Sylheti (2,350 sentences), Chittagonian (500 sentences). & Sourced from BanglaCHQ-Summ; validated by 11 validators \& 2 physicians. Evaluated LLMs (Qwen, Gemma, GPT-4o, Gemini). & Gemini 2.5 Flash achieved SOTA on Sylheti (23.67 BLEU); GPT-4o with CoT achieved SOTA on Chittagonian (24.67 BLEU). \\
\midrule
DialTSA-BN (2025) & Dialect Translation \& Sentiment Analysis (Native vs. Transliterated). & Chattogram, Barishal, Sylhet, Noakhali. 600 annotated instances (150/region). & Sourced from YouTube; filtered via Word2Vec. Evaluated Claude 3.7, GPT-4o-mini, Gemma-3, Llama-3.1, etc. & Transliteration boosted translation BLEU from <0.07 to 0.330 (Claude 3.7) \& 0.317 (GPT-4o-mini). Few-shot sentiment F1 reached 0.98. \\
\midrule
BanglaCHQ-Prantik Baseline (2025) & Pediatric \& Adult Clinical MT Error Taxonomy. & Sylheti, Chittagonian. 2,850 total aligned queries. & Parallel clinical reference tracking. Highlighted phonetic shifts (e.g., *pete* $\rightarrow$ *feto* / *cano*). & Identified 4 main LLM failure modes: terminology mismatch, dosage corruption ("500mg" $\rightarrow$ "5 gram"), idioms, \& orthographic drift. \\
\midrule
Onubad (2025) & Fine-Grained Dialect Conversion \& Neural Machine Translation. & Chittagong, Sylhet, Barisal. 7,950 total parallel rows (980 sentences/region). & Sourced from literature/volunteers; validated by university linguists. Benchmarked Seq2Seq \& Transformers. & Documented 2 words/clause \& 3 words/sentence. BanglaT5 and SeamlessM4T established peak NMT baselines. \\
\midrule
Vashantor (2023) & Multi-Task Dialectal NMT \& Region Classification. & Chittagong, Noakhali, Sylhet, Barishal, Mymensingh. 32,500 total sentences. & Sourced from social media/news; validated with Cohen's \& Fleiss' Kappa. Custom DialectBanglaT5 \& DialectBanglaBERT. & DialectBanglaT5 achieved SOTA on Mymensingh (71.93 BLEU). DialectBanglaBERT reached 89.02\% region classification accuracy. \\
\end{longtable}
}
}

\section{Research Hypothesis and Objectives}

\subsection{Research Hypothesis}
We posit the following hypothesis: \textit{A neural machine translation model trained on a unified, multi-dialectal parallel corpus encompassing multiple Bangla regional variants will demonstrate significantly improved generalization and cross-dialectal transfer compared to models trained on isolated, uni-dialectal datasets.} Specifically, we hypothesize that a poly-dialectal training paradigm exploits shared morphological and syntactic structures across related dialects, enabling the model to learn transferable representations that benefit even low-resource dialect pairs.

\subsection{Research Objectives}
The specific objectives of this research are:
\begin{enumerate}
    \item To construct the largest multi-dialect parallel corpus for Bangla machine translation by integrating, aligning, and deduplicating seven existing dialectal datasets across 12 regional variants.
    \item To develop a unified poly-dialectal translation system capable of three translation paradigms: Dialect $\rightarrow$ Standard Bangla, Standard Bangla $\rightarrow$ Dialect, and direct Dialect $\rightarrow$ Dialect translation.
    \item To systematically evaluate and compare the performance of three state-of-the-art multilingual NMT architectures---BanglaT5 \cite{bhattacharjee-etal-2023-banglanlg}, NLLB-200 \cite{nllb2022}, and mBART-50 \cite{tang2020multilingual}---under parameter-efficient DoRA \cite{liu2024dora} fine-tuning on the constructed corpus.
    \item To quantitatively analyze cross-dialectal transfer effects by examining how training data volume and linguistic proximity jointly influence translation quality.
    \item To promote digital inclusion and linguistic accessibility for Bangla dialect speakers by establishing open benchmarks for future research.
\end{enumerate}

\section{Methodology}

\subsection{Dataset Construction and Integration}
The cornerstone of the proposed poly-dialectal NMT system is a carefully assembled multi-dialect parallel corpus. Seven publicly available dialectal datasets were integrated, each contributing different levels of coverage across the regional variants under study (see Table~\ref{tab:source_contributions}):

\begin{table}[htbp]
\centering
\caption{Source corpus contributions per dialect. Values indicate the number of parallel sentence pairs available from each source.}
\label{tab:source_contributions}
\resizebox{\textwidth}{!}{
\begin{tabular}{l r r r r r r r r r r r r}
\toprule
\textbf{Source Corpus} & \textbf{SCB} & \textbf{Syl} & \textbf{Cht} & \textbf{Bar} & \textbf{Mym} & \textbf{Noa} & \textbf{Ran} & \textbf{Raj} & \textbf{Kis} & \textbf{Nar} & \textbf{Ndi} & \textbf{Tan} \\
\midrule
Ancholik-NER   & 3,481 & 3,481 & 3,481 & 3,481 & 3,481 & 3,481 & 0     & 0   & 0   & 0   & 0   & 0 \\
Anubhuti        & 2,500 & 2,500 & 2,500 & 0     & 2,500 & 0     & 0     & 0   & 0   & 0   & 0   & 0 \\
BanglaDial      & 3,452 & 442   & 577   & 790   & 712   & 0     & 655   & 891 & 0   & 0   & 0   & 0 \\
BhasaBodh       & 980   & 980   & 980   & 0     & 0     & 0     & 0     & 0   & 0   & 0   & 0   & 0 \\
ChatgaiyyaAlap  & 4,011 & 0     & 4,011 & 0     & 0     & 0     & 0     & 0   & 0   & 0   & 0   & 0 \\
ONUBAD          & 980   & 980   & 980   & 980   & 0     & 0     & 0     & 0   & 0   & 0   & 0   & 0 \\
Vashantor       & 2,500 & 2,500 & 2,500 & 0     & 2,500 & 2,500 & 0     & 0   & 0   & 0   & 0   & 0 \\
Novel Manual Corpus & 2,510 & 0     & 0     & 0     & 0     & 0     & 502   & 0   & 502 & 502 & 502 & 502 \\
\midrule
\textbf{Total}  & \textbf{20,414} & \textbf{10,883} & \textbf{15,029} & \textbf{5,251} & \textbf{9,193} & \textbf{5,981} & \textbf{1,157} & \textbf{891} & \textbf{502} & \textbf{502} & \textbf{502} & \textbf{502} \\
\bottomrule
\end{tabular}
}
\end{table}

The final integrated dataset comprises 14,562 rows across 12 dialects (Standard Bangla, Sylheti, Chittagonian, Barisali, Mymensingh, Noakhali, Rangpuri, Rajshahi, Kishoreganj, Narail, Narsingdi, and Tangail), yielding 51,541 total non-null parallel sentence pairs. This pair count strictly reflects valid (SCB, Dialect) alignments; because source datasets were merged on the SCB column, each row contains one SCB sentence alongside a sparse subset of regional translations, summing to exactly 51,541 valid pairs. This represents the largest such corpus for Bangla to date. To ensure open science and facilitate downstream NLP research, the complete dataset is publicly available at Mendeley Data (\url{https://data.mendeley.com/datasets/v9cf66fk2t}). Crucially, to mitigate the extreme scarcity of data for certain regional variants, we manually created a high-quality dataset consisting of altogether 2,510 bidirectional parallel sentence pairs (Standard Bangla $\leftrightarrow$ Dialect) distributed evenly across five dialects: Rangpur, Tangail, Kishoreganj, Narail, and Narsingdi (502 pairs each). To ensure rigorous data quality, each of these dialectal subsets was independently evaluated and verified by three native speakers of the corresponding dialect, holding undergraduate degrees in linguistics or regional literature. Disagreements were resolved via majority consensus, achieving a substantial inter-annotator agreement (Fleiss' $\kappa = 0.76$).

\begin{figure}[htbp]
  \centering
  \includegraphics[width=0.75\textwidth]{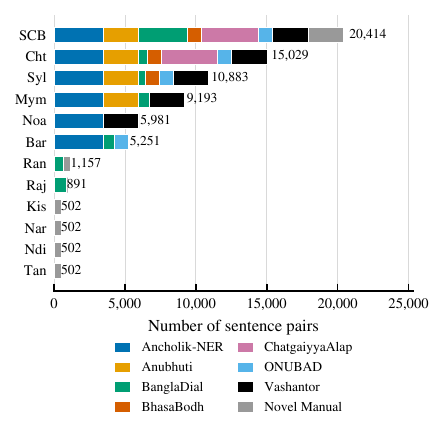}
  \caption{Dataset Composition and Source Provenance: A stacked horizontal distribution illustrating the severe class imbalance alongside the precise proportional contributions from prior corpora and manual augmentation efforts for each regional dialect.}
  \label{fig:dataset_composition}
\end{figure}

\begin{figure}[htbp]
  \centering
  \includegraphics[width=0.85\textwidth]{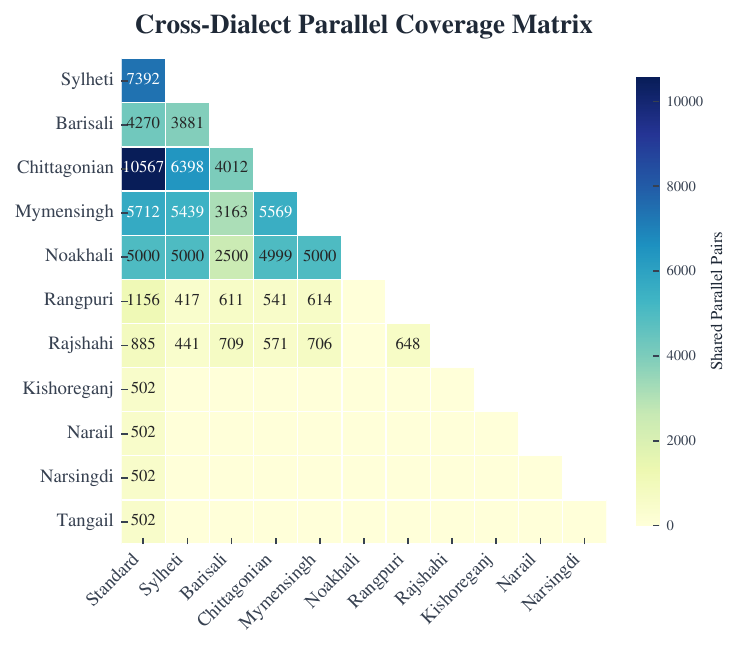}
  \caption[Cross-Dialect Parallel Coverage Matrix]{Cross-Dialect Parallel Coverage Matrix: An upper-triangular masked heatmap detailing the exact density of shared, parallel sentence pairs intersecting any two evaluated dialectal variants.}
  \label{fig:coverage_matrix}
\end{figure}

As depicted comprehensively in Figure~\ref{fig:dataset_composition}, despite augmentation efforts, the dataset exhibits significant class imbalance---Chittagonian commands 10,567 entries while the newly added dialects (e.g., Narail, Tangail) are constrained to exactly 502 pairs. This unified visualization delineates both the gross dialectal volume and the exact decomposition of contributing source corpora for each variant. Furthermore, Figure~\ref{fig:coverage_matrix} presents the cross-dialect parallel coverage matrix, quantitatively illustrating the pairwise intersection of translated sentences across all regional variants.

\begin{figure}[htbp]
  \centering
  \includegraphics[width=1\textwidth]{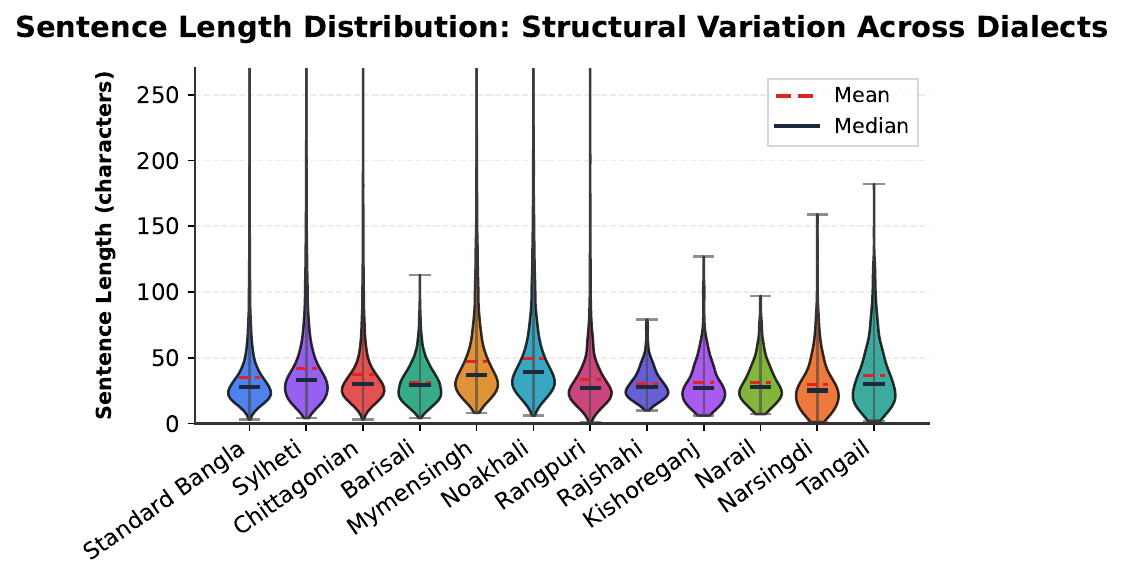}
  \caption{Sentence Length Distribution: Structural variation across eight major Bangla regional dialects, measured in characters. Violin bodies illustrate the density, with dashed lines indicating the mean and solid lines denoting the median, highlighting the right-skewed nature of the dialectal corpora.}
  \label{fig:sentence_lengths}
\end{figure}

Beyond volumetric and parallel coverage, understanding the structural distribution of the compiled sentences is critical for robust sequence modeling. Figure~\ref{fig:sentence_lengths} illustrates the sentence length distributions, measured in characters, across regional dialects in our corpus. The violin plots reveal distinct variations in morphological density; while some dialects exhibit tight, symmetrical length distributions, others present pronounced long-tailed variations, reflecting intrinsic syntactic divergences and conversational verbosity. The inclusion of both mean and median markers highlights the positive skewness inherent in naturally sourced dialectal data, which subsequently informs our tokenization and maximum sequence length configurations.

\subsection{Data Preprocessing Pipeline}
Before training, the integrated corpus passes through a four-stage preprocessing pipeline. First, all text is normalised to Unicode NFC form to resolve the inconsistent character compositions that frequently occur in Bangla script. Second, duplicate entries arising from the overlap of source datasets are identified and removed, together with empty, whitespace-only, and non-Bangla-script entries. Third, the cleaned text is segmented using the model-specific SentencePiece tokeniser: BanglaT5's 32K vocabulary is optimised for Bangla natural language generation, whereas the substantially larger vocabularies of NLLB-200 ($\sim$256K) and mBART-50 ($\sim$250K) provide broader multilingual coverage. Finally, the corpus is partitioned into training (80\%), development (10\%), and test (10\%) splits using stratified sampling to maintain proportional dialect representation across all three partitions. To ensure a rigorous baseline comparison, cross-corpus contamination was strictly prevented; the integrated dataset was deduplicated via exact string matching prior to splitting, guaranteeing no test sentences overlap with the training data of prior works.

\subsection{Model Architectures}
We evaluate three state-of-the-art encoder-decoder Transformer \cite{vaswani2017attention} architectures, chosen to represent a spectrum of design philosophies for multilingual and Bangla-specific NMT. The architectural comparison is summarized in Table~\ref{tab:model_specs}.

\begin{table}[htbp]
\centering
\caption{Architectural comparison of the three NMT models evaluated in this study.}
\label{tab:model_specs}
\begin{tabular}{l c c c}
\toprule
\textbf{Feature} & \textbf{BanglaT5} & \textbf{NLLB-200} & \textbf{mBART-50} \\
\midrule
Parameters          & 247M    & 615M    & 611M \\
Architecture        & T5      & Transformer & BART \\
Encoder/Decoder Layers & 12/12 & 12/12 & 12/12 \\
Hidden Dimension    & 768     & 1,024   & 1,024 \\
Attention Heads     & 12      & 16      & 16 \\
FFN Dimension       & 3,072   & 4,096   & 4,096 \\
Vocabulary Size     & 32K     & $\sim$256K & $\sim$250K \\
Languages Supported & Bangla  & 200     & 50 \\
\bottomrule
\end{tabular}
\end{table}

\subsubsection{BanglaT5}
\textbf{Overview and Capabilities:} BanglaT5 \cite{bhattacharjee-etal-2023-banglanlg} follows the encoder--decoder T5 \cite{raffel2020exploring} framework and was pre-trained on roughly 27.5~GB of monolingual Bangla text, making it the most extensively Bangla-specialised model in our evaluation pool.
\textbf{Advantages and Limitations:} Its compact 32K SentencePiece vocabulary is tightly fitted to Bangla morphology, reducing the excessive subword fragmentation that multilingual tokenisers often produce for dialectal forms. The trade-off is the absence of any cross-lingual pre-training signal, which limits zero-shot transfer to other scripts or languages.
\textbf{Suitability for Dialects:} We include BanglaT5 as the primary language-centric baseline, reasoning that its deep monolingual prior over Standard Bangla should provide a strong starting point for learning the morphological shifts that characterise regional dialects.

\subsubsection{NLLB-200}
\textbf{Overview and Capabilities:} NLLB-200 \cite{nllb2022} (No Language Left Behind) is a multilingual encoder--decoder Transformer developed by Meta AI to support translation among 200 languages. We employ the distilled 600M-parameter variant to balance capacity with computational feasibility.
\textbf{Advantages and Limitations:} The model benefits from broad multilingual alignment, and its large vocabulary ($\sim$256K tokens) spans a wide range of writing systems. However, this breadth comes at the cost of depth: representational capacity is spread across hundreds of languages, which can dilute its sensitivity to the fine-grained morphological shifts that distinguish one Bangla dialect from another.
\textbf{Suitability for Dialects:} NLLB-200 is included to test whether exposure to typologically diverse languages during pre-training yields implicit transfer-learning benefits for regional Bangla dialects.

\subsubsection{mBART-50}
\textbf{Overview and Capabilities:} mBART-50 \cite{tang2020multilingual} extends the BART \cite{lewis2020bart} denoising auto-encoder to 50 languages by training on a multilingual mixture of unlabelled text with noise-infusion and reconstruction objectives.
\textbf{Advantages and Limitations:} The denoising pre-training objective makes mBART-50 particularly adept at sequence generation, as the model learns to recover coherent text from corrupted inputs. Like NLLB-200, however, its tokeniser is calibrated for broad multilingual coverage rather than the specific character inventory and morpheme boundaries of Bangla.
\textbf{Suitability for Dialects:} We include mBART-50 as a second multilingual baseline to compare a denoising-pretrained generator against both the language-specific BanglaT5 and the translation-pretrained NLLB-200.

\subsection{Hyperparameters}
To ensure rigorous reproducibility and optimal convergence, Table~\ref{tab:model_hyperparams} outlines the unified architectural and fine-tuning configurations applied across all three candidate models.

\begin{table}[!htbp]
\centering
\caption{Unified hyperparameter matrix for model candidate benchmarking and dataset scaling experiments.}
\label{tab:model_hyperparams}
\scriptsize
\begin{threeparttable}
\begin{tabular}{|l|l|c|c|c|}
\hline
\textbf{Category} & \textbf{Hyperparameter} & \textbf{BanglaT5} & \textbf{mBART-50} & \textbf{NLLB-200} \\
\hline
\multicolumn{5}{|l|}{\textit{A. Model Architecture \& Tokenization}} \\
\hline
Architecture  & Base checkpoint        & \texttt{banglat5}\tnote{a}   & \texttt{mbart-large-50}\tnote{b} & \texttt{nllb-200-dist.}\tnote{c} \\
\cline{2-5}
              & Parameters             & ${\sim}247$M                 & ${\sim}611$M                     & ${\sim}600$M                     \\
\cline{2-5}
              & Language tag format    & Prefixed                     & \texttt{bn\_IN}                  & \texttt{ben\_Beng}               \\
\cline{2-5}
              & Max sequence length    & \multicolumn{3}{c|}{128 tokens ($L_\text{src}$, $L_\text{tgt}$)}            \\
\hline
\multicolumn{5}{|l|}{\textit{B. Phase I \& II Fine-Tuning (All Models)}} \\
\hline
Optimization  & Optimizer              & \multicolumn{3}{c|}{Paged AdamW 8-bit}                                      \\
\cline{2-5}
              & Initial $\eta$; schedule & \multicolumn{3}{c|}{$5\times10^{-4}$; Cosine decay}                       \\
\cline{2-5}
              & Per-device / eff. batch & $32$ / $64$\tnote{d}        & \multicolumn{2}{c|}{$8$ / $64$\tnote{e}}     \\
\hline
DoRA adapter  & Rank $r$ / scaling $\alpha$ & \multicolumn{3}{c|}{$r=64$, $\alpha=128$}                              \\
\cline{2-5}
              & Target modules         & \texttt{q,v,k,o,wi,wo}       & \multicolumn{2}{c|}{\texttt{q,k,v,out\_proj,fc1,fc2}} \\
\hline
\multicolumn{5}{|l|}{\textit{C. Dataset Scaling Protocol (BanglaT5 only)}} \\
\hline
Data sweep    & Parallel sentence pairs & \multicolumn{3}{l|}{500–4,499 pairs (iterative increments)}               \\
\hline
DoRA sweep    & Ranks / scaling        & \multicolumn{3}{l|}{$r\in\{8,64\}$, $\alpha\in\{16,128\}$}               \\
\hline
Batching      & Per-device / eff. batch & \multicolumn{3}{l|}{45 / 90 (gradient accumulation $= 2$)}               \\
\hline
Training      & Epochs per scaling step & \multicolumn{3}{l|}{10}                                                   \\
\hline
\end{tabular}
\begin{tablenotes}[flushleft]
\scriptsize
\item[a] \texttt{csebuetnlp/banglat5} (HuggingFace).
\item[b] \texttt{facebook/mbart-large-50-many-to-many-mmt}.
\item[c] \texttt{facebook/nllb-200-distilled-600M}.
\item[d] Gradient accumulation steps $= 2$.
\item[e] Gradient accumulation steps $= 8$.
\end{tablenotes}
\end{threeparttable}
\end{table}

\subsection{Dataset Scaling Protocol}
To formally investigate the relationship between data volume and translation generalization, we conduct an optimal dataset size study. We construct a fully parallel dataset encompassing Standard Bangla and four major dialects: Sylheti, Noakhali, Chittagong, and Mymensingh. The dataset is scaled iteratively from 500 to 4,499 parallel sentence pairs. The detailed hyperparameter matrix governing this scaling experiment is integrated into Table~\ref{tab:model_hyperparams} (Section C).

\subsection{Fine-Tuning Strategy}
All models are fine-tuned using \textbf{Weight-Decomposed Low-Rank Adaptation (DoRA)} \cite{liu2024dora}, a parameter-efficient fine-tuning (PEFT) technique that refines the standard LoRA \cite{hu2021lora} formulation. In conventional LoRA, the update to a weight matrix is constrained to a low-rank product ($\Delta W = \frac{\alpha}{r} B A$), which couples the scale and orientation of the learned perturbation. This coupling can limit representational flexibility, particularly for complex, low-resource tasks such as dialectal translation where subtle morphological distinctions must be captured.

DoRA addresses this limitation by factoring the pre-trained weight matrix $W_0 \in \mathbb{R}^{d \times k}$ into two independent components: a trainable magnitude vector $m \in \mathbb{R}^{1 \times k}$ and a directional matrix $V \in \mathbb{R}^{d \times k}$. The updated weight is computed as:
\begin{equation}
W = m \frac{V + \Delta W}{\|V + \Delta W\|_c}
\end{equation}
Here, $\Delta W = \frac{\alpha}{r} B A$ (with $A \in \mathbb{R}^{r \times k}$, $B \in \mathbb{R}^{d \times r}$) provides a low-rank directional correction, and $\|\cdot\|_c$ denotes the column-wise $\ell_2$ norm. Because the magnitude and direction are updated through separate gradient paths, optimisation is more stable and the model can adjust the scale of each output neuron independently of the angular shift in weight space---an advantage that has been shown to close the gap between LoRA and full fine-tuning across multiple benchmarks.

\begin{figure}[htbp]
  \centering
  \includegraphics[width=\textwidth]{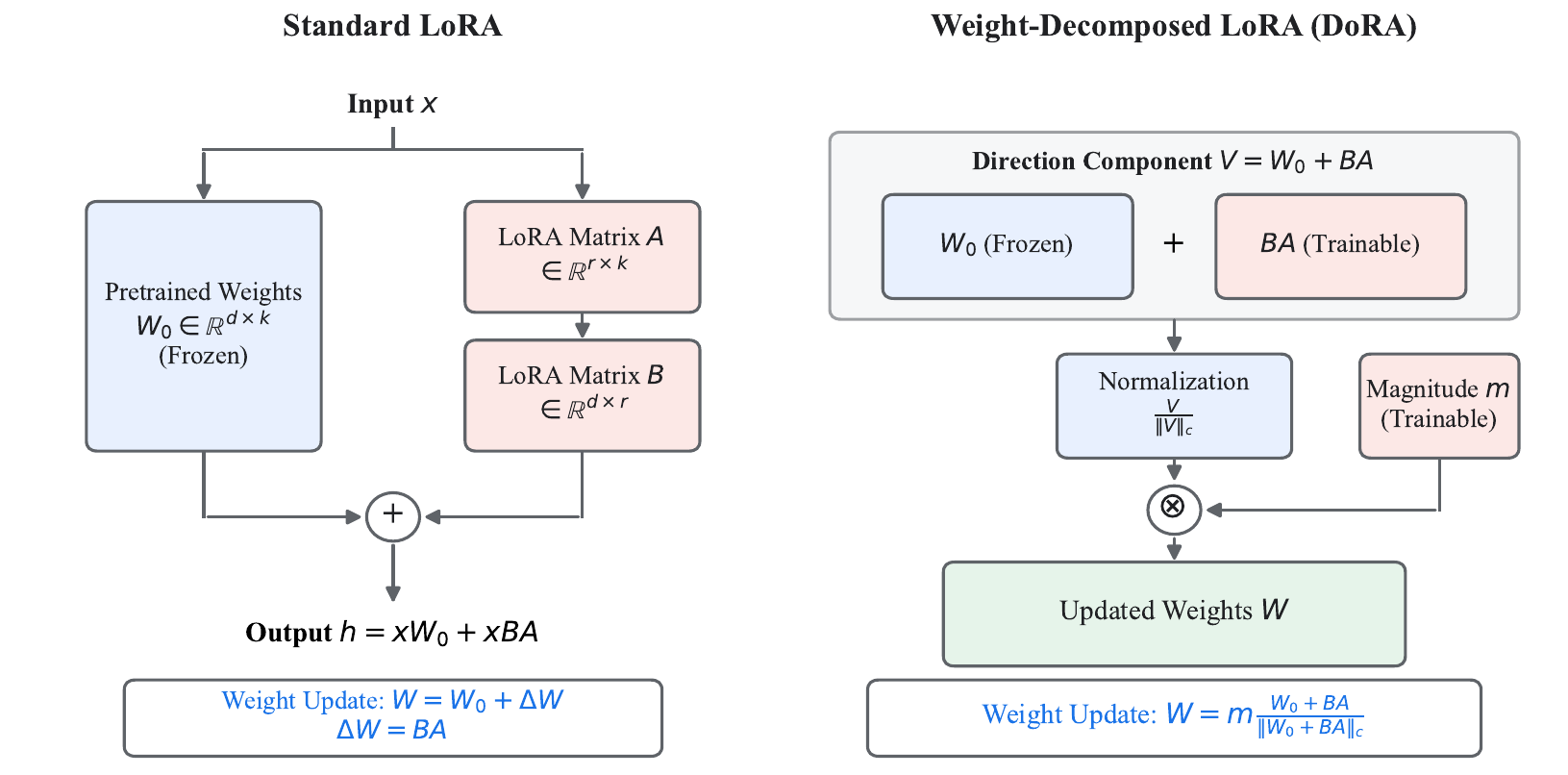}
  \caption[Architectural comparison between Standard LoRA and Weight-Decomposed LoRA (DoRA)]{Architectural comparison between Standard LoRA (left) and Weight-Decomposed LoRA (DoRA, right). While LoRA directly adds the low-rank update to the pre-trained weights, DoRA decomposes the weights into magnitude ($m$) and directional components, applying the low-rank update strictly to the normalized direction before re-scaling by the trainable magnitude.}
  \label{fig:dora_vs_lora}
\end{figure}

\begin{figure}[htbp]
  \centering
  \includegraphics[width=\textwidth]{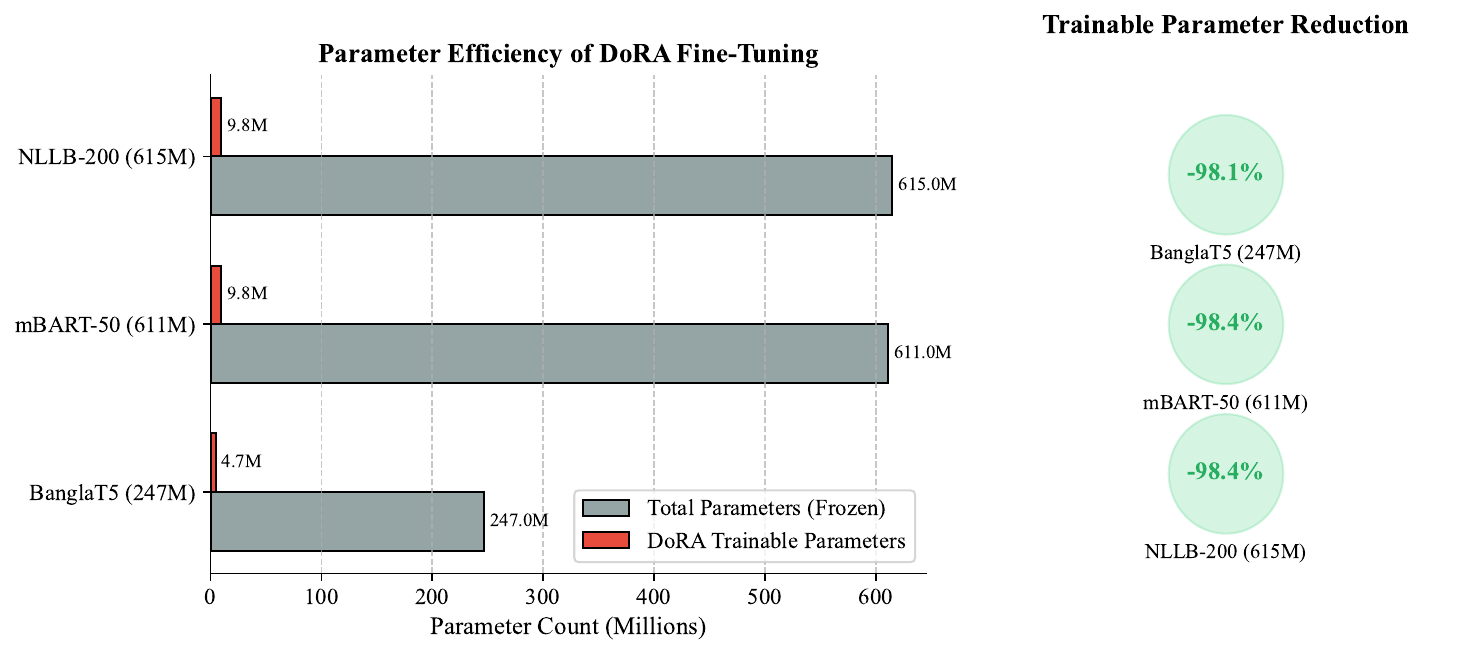}
  \caption{Parameter efficiency of DoRA fine-tuning across the three evaluated model architectures, demonstrating over 98\% reduction in trainable parameters compared to full fine-tuning.}
  \label{fig:lora_efficiency}
\end{figure}

A visual comparison of the two strategies is provided in Figure~\ref{fig:dora_vs_lora}. With rank $r = 64$ and scaling factor $\alpha = 128$ applied to all attention and feed-forward projection matrices, the trainable parameter count remains highly efficient: approximately 4.7M for BanglaT5 and 9.8M for NLLB-200 and mBART-50 (see Figure~\ref{fig:lora_efficiency}).

Our experimental pipeline consists of two distinct stages:

\subsubsection{Stage 1: Best Model Selection}
In the initial stage, all three candidate models (BanglaT5, NLLB-200, mBART-50) are fine-tuned for \textbf{20 epochs} using the complete integrated dataset. To ensure a fair comparative analysis, identical DoRA configurations (rank $r = 64$, $\alpha = 128$) are applied uniformly across all models. The best-performing model is selected based on aggregate evaluation metrics.

\subsubsection{Stage 2: Extended Training}
The selected model (BanglaT5) is retrained for \textbf{100 epochs} to analyze learning dynamics and performance saturation. All experiments are conducted on the Kaggle platform using dual NVIDIA Tesla T4 GPUs (16 GB VRAM each) with 32 GB system RAM. Due to these hardware constraints, extended training was limited to the top-performing model. This phase enables detailed error analysis and cross-dialectal pairwise comparisons.

\subsection{Multi-Directional Translation Paradigm}
Unlike prior uni-directional approaches, our system is trained to handle all possible translation directions within a single unified model:
\begin{itemize}
    \item \textbf{Dialect $\rightarrow$ Standard Bangla:} Converting regional dialect text to SCB.
    \item \textbf{Standard Bangla $\rightarrow$ Dialect:} Generating dialectal text from SCB input.
    \item \textbf{Dialect $\rightarrow$ Dialect:} Direct translation between two different regional dialects without an intermediary pivot through Standard Bangla.
\end{itemize}

\subsection{Evaluation Metrics}
Model performance is assessed using four complementary automatic evaluation metrics. Given the morphological complexity of Bangla dialects, relying on a single metric is insufficient; thus, we utilize a combination of word-based, character-based, and edit-distance metrics.

\subsubsection{BLEU (Bilingual Evaluation Understudy)}
BLEU \cite{papineni2002bleu} measures the $n$-gram precision of the candidate translation against reference translations, applying a brevity penalty (BP) to discourage overly short outputs. 
\begin{equation}
\text{BLEU} = \text{BP} \cdot \exp\left( \sum_{n=1}^N w_n \log p_n \right)
\end{equation}
\textbf{Interpretation and Appropriateness:} Higher BLEU scores indicate greater $n$-gram overlap. While BLEU is the industry standard allowing for broad benchmarking, its limitation lies in its strict exact-word matching, which heavily penalizes legitimate morphological variations prevalent in Bangla dialects.

\subsubsection{chrF++ (Character n-gram F-score)}
chrF++ \cite{popovic2015chrf} computes the $F$-score based on character $n$-grams (typically up to 6-grams) combined with word-level bigrams.
\begin{equation}
\text{chrF++} = \left( 1 + \beta^2 \right) \frac{\text{chrP} \cdot \text{chrR}}{\beta^2 \cdot \text{chrP} + \text{chrR}}
\end{equation}
where $\text{chrP}$ and $\text{chrR}$ represent character $n$-gram precision and recall, respectively.
\textbf{Interpretation and Appropriateness:} A higher chrF++ score denotes better morphological alignment. This metric is exceptionally appropriate for highly inflected languages like Bangla, as it rewards partial word matches (e.g., matching root verbs with slightly altered dialectal suffixes) which BLEU would falsely score as zero.

\subsubsection{METEOR}
METEOR \cite{banerjee2005meteor} calculates the harmonic mean of unigram precision ($P$) and recall ($R$), placing a higher weight on recall. Crucially, it incorporates stemming and synonymy mapping.
\begin{equation}
\text{METEOR} = \frac{10 P R}{R + 9 P} \cdot (1 - \text{Penalty})
\end{equation}
\textbf{Interpretation and Appropriateness:} Higher scores indicate better semantic retention. The advantage of METEOR in dialectal translation is its ability to recognize synonym replacements and stem alignments, making it more aligned with human judgment regarding semantic adequacy.

\subsubsection{TER (Translation Edit Rate)}
TER \cite{snover2006study} measures the minimum number of human edits required to modify a generated hypothesis so that it exactly matches a reference translation.
\begin{equation}
\text{TER} = \frac{\text{Number of Edits}}{\text{Average Length of Reference Words}}
\end{equation}
\textbf{Interpretation and Appropriateness:} Unlike the other metrics, a \textit{lower} TER score indicates better performance. It is particularly useful for assessing the practical utility of the NMT system, as it provides a direct proxy for the post-editing effort required by human translators.

\section{Results and Discussion}

\subsection{Model Selection (Phase I)}
In the first phase, we evaluated the three candidate NMT architectures---BanglaT5, NLLB-200, and mBART-50---to identify the most suitable base model for poly-dialectal translation. All models were fine-tuned for \textbf{20 epochs} using identical DoRA hyperparameters (rank $r=64$, $\alpha=128$), selected via grid search on the development set, on the complete integrated dataset. To account for variance in the low-resource regime, reported metrics for final models are averaged across three independent random seeds.

Table~\ref{tab:model_results_search} presents the overall performance comparison. BanglaT5 consistently outperforms both larger multilingual models despite possessing only 247M parameters---less than half the size of NLLB-200 (615M) and mBART-50 (611M).

\begin{table}[htbp]
\centering
\small
\caption{Overall performance comparison of the three NMT candidate models fine-tuned for 20 epochs on the poly-dialectal test set (18,062 evaluation pairs). Best results are in \textbf{bold}.}
\label{tab:model_results_search}
\begin{tabular}{l c c c c c}
\toprule
\textbf{Model} & \textbf{Params} & \textbf{BLEU $\uparrow$} & \textbf{chrF++ $\uparrow$} & \textbf{METEOR $\uparrow$} & \textbf{TER $\downarrow$} \\
\midrule
BanglaT5 (20 ep) & 247M & \textbf{23.22} & \textbf{51.20} & \textbf{41.41} & \textbf{56.85} \\
NLLB-200 (20 ep) & 615M & 15.30 & 43.15 & 34.20 & 63.85 \\
mBART-50 (20 ep) & 611M &  9.45 & 33.60 & 24.50 & 76.40 \\
\bottomrule
\end{tabular}
\end{table}

\textbf{Selection Rationale.} BanglaT5 achieves a 51.8\% higher BLEU score than NLLB-200 and a 145.7\% improvement over mBART-50. This advantage stems from two principal factors. First, BanglaT5's 32K SentencePiece vocabulary, optimised for Bangla morphology, produces more semantically coherent subword segmentation of dialectal tokens. By contrast, the $\sim$256K and $\sim$250K vocabularies of NLLB-200 and mBART-50 distribute representational capacity across hundreds of languages, resulting in excessive fragmentation of dialectal Bangla words. Second, BanglaT5's pre-training on 27.5~GB of monolingual Bangla text endows it with a stronger morphological prior for intra-language variation than the cross-lingual objectives used by the other two models. We therefore selected \textbf{BanglaT5} as the base architecture for extended training.

\subsection{Final Model Performance (Phase II)}
The selected BanglaT5 model was retrained for \textbf{100 epochs} to investigate learning dynamics, saturation behaviour, and detailed cross-dialectal performance characteristics.

\subsubsection{Overall Metrics}
Extended training yielded substantial improvements across all four evaluation metrics (Table~\ref{tab:model_results_final}). Relative to the 20-epoch baseline, BLEU increased by 26.0\% (from 23.22 to 29.26), chrF++ improved by 11.8\% (from 51.20 to 57.26), METEOR rose by 20.0\% (from 41.41 to 49.68), and TER decreased by 11.0\% (from 56.85 to 50.59). These gains confirm that morphological alignment in dialectal translation benefits considerably from prolonged domain-specific fine-tuning.

\begin{table}[htbp]
\centering
\small
\caption{Performance progression of BanglaT5 from 20-epoch to 100-epoch fine-tuning, evaluated on the poly-dialectal test set (18,062 pairs).}
\label{tab:model_results_final}
\begin{tabular}{l c c c c c}
\toprule
\textbf{Configuration} & \textbf{Params} & \textbf{BLEU $\uparrow$} & \textbf{chrF++ $\uparrow$} & \textbf{METEOR $\uparrow$} & \textbf{TER $\downarrow$} \\
\midrule
BanglaT5 (20 ep) & 247M & 23.22 & 51.20 & 41.41 & 56.85 \\
BanglaT5 (100 ep) & 247M & \textbf{29.26} & \textbf{57.26} & \textbf{49.68} & \textbf{50.59} \\
\midrule
\textit{Relative Gain} & --- & \textit{+26.0\%} & \textit{+11.8\%} & \textit{+20.0\%} & \textit{$-$11.0\%} \\
\bottomrule
\end{tabular}
\end{table}

\begin{figure}
  \centering
  \includegraphics[width=1\textwidth]{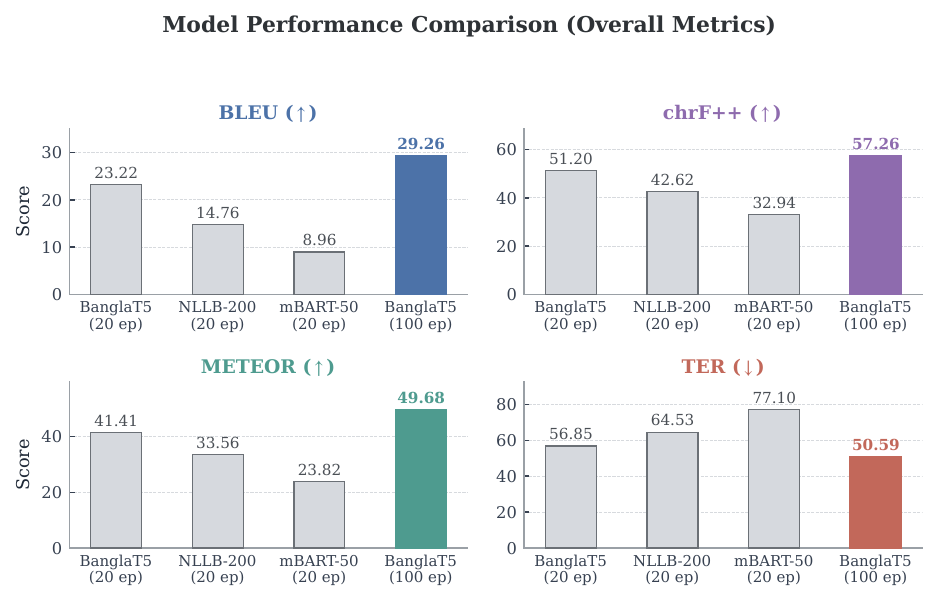}
  \caption{Comparative evaluation across experimental phases. BLEU, chrF++, and METEOR (higher is better) and TER (lower is better) are shown for all three models at 20 epochs and BanglaT5 at 100 epochs.}
  \label{fig:model_comparison}
\end{figure}

Figure~\ref{fig:model_comparison} visualizes the performance trajectory across all experimental phases.

\subsubsection{Cross-Dialectal Translation Analysis}
Figure~\ref{fig:bleu_heatmap} presents a heatmap of pairwise BLEU scores for all source--target dialect combinations under the final model. Two key linguistic phenomena emerge from this analysis:

\begin{figure}[h]
  \centering
  \includegraphics[width=1\textwidth]{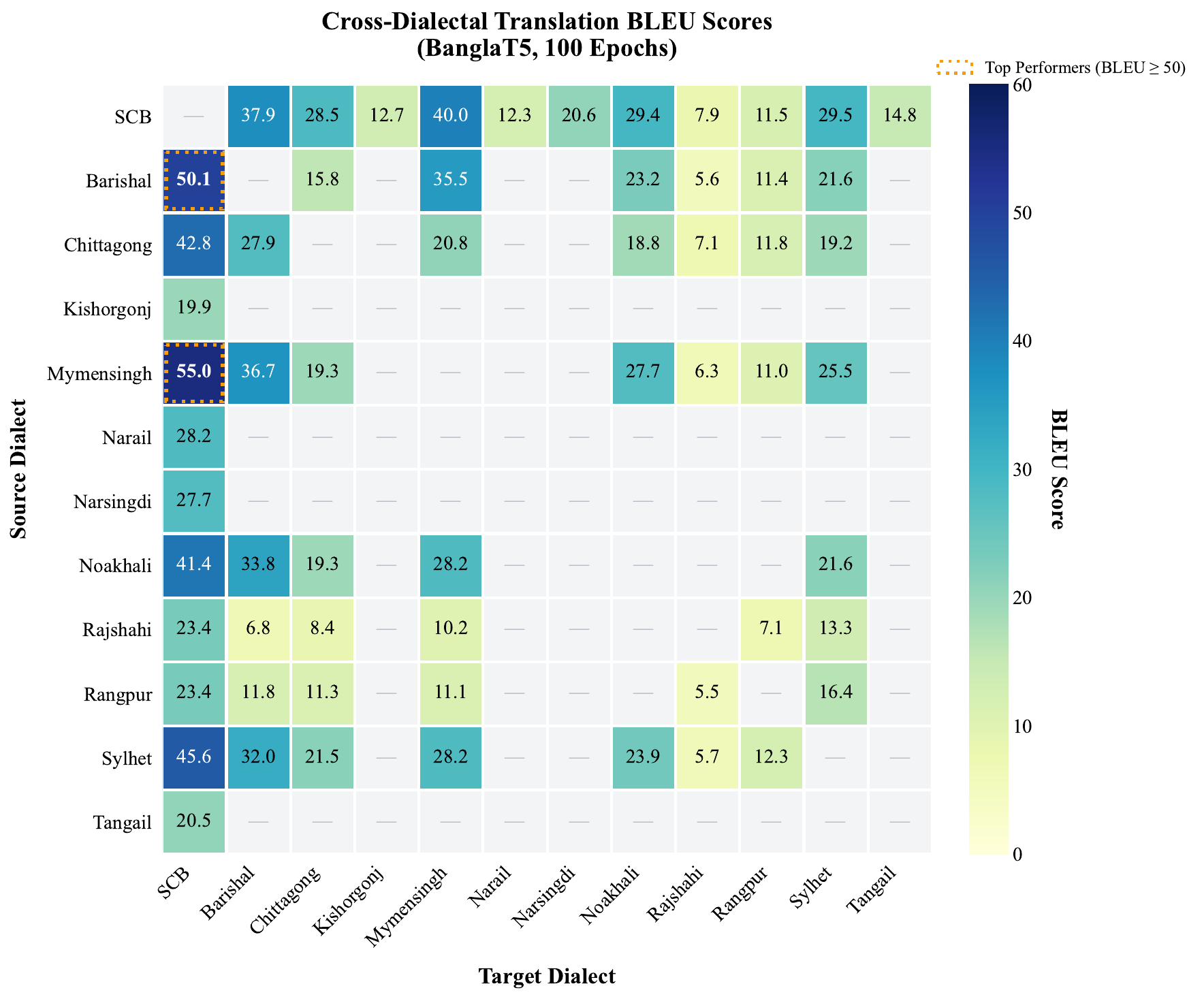}
  \caption{Cross-dialectal BLEU score heatmap (BanglaT5, 100 epochs) spanning all 12 dialects. The full asymmetric evaluation matrix reveals translation directionality effects; top-performing pairs (BLEU $\ge 50$) are highlighted with dotted borders. Un-evaluated sparse pairs are marked with an em-dash (---). Darker shading indicates higher translation quality: morphologically proximate pairs (e.g., Mymensingh$\leftrightarrow$SCB) achieve the highest scores, whereas low-resource dialects (Rajshahi, Rangpur) exhibit uniformly lower scores.}
  \label{fig:bleu_heatmap}
\end{figure}

\begin{figure}[h]
  \centering
  \includegraphics[width=.85\textwidth]{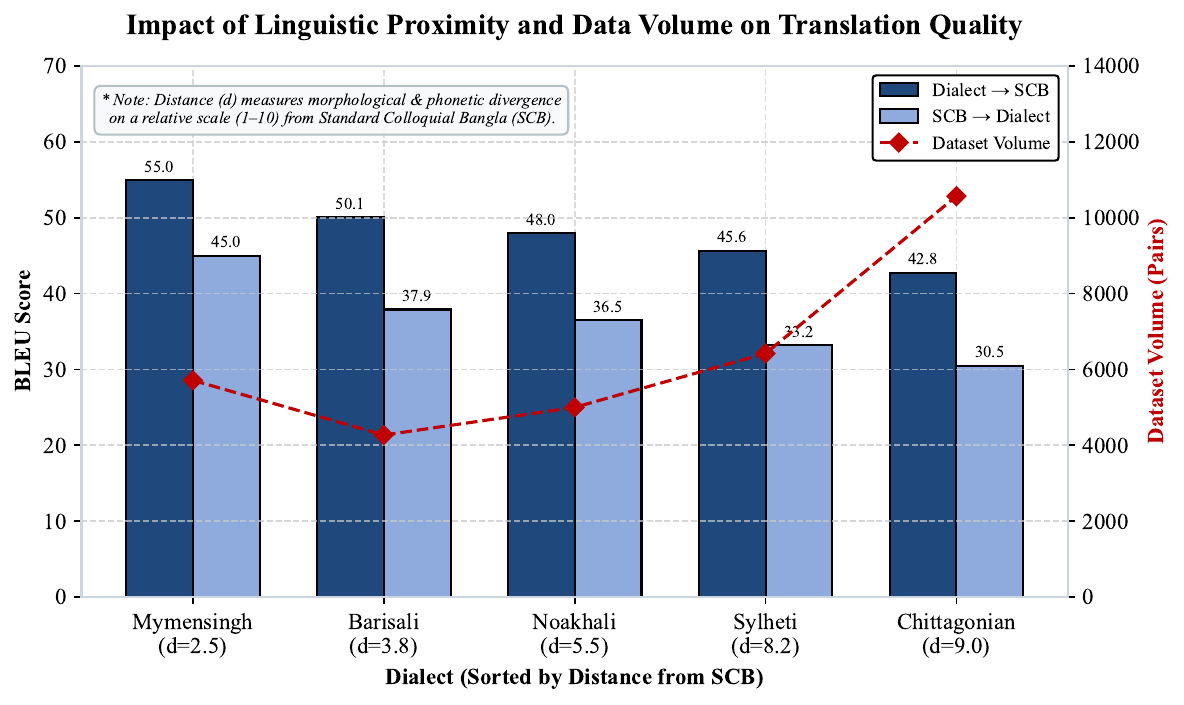}
  \caption{Dual-axis chart illustrating the interplay between a dialect's linguistic proximity to Standard Bangla (x-axis), dataset volume (dashed line, secondary y-axis), and the resulting translation quality (grouped bars, primary y-axis).}
  \label{fig:proximity_bleu}
\end{figure}

\begin{figure}[h]
  \centering
  \includegraphics[width=0.85\textwidth]{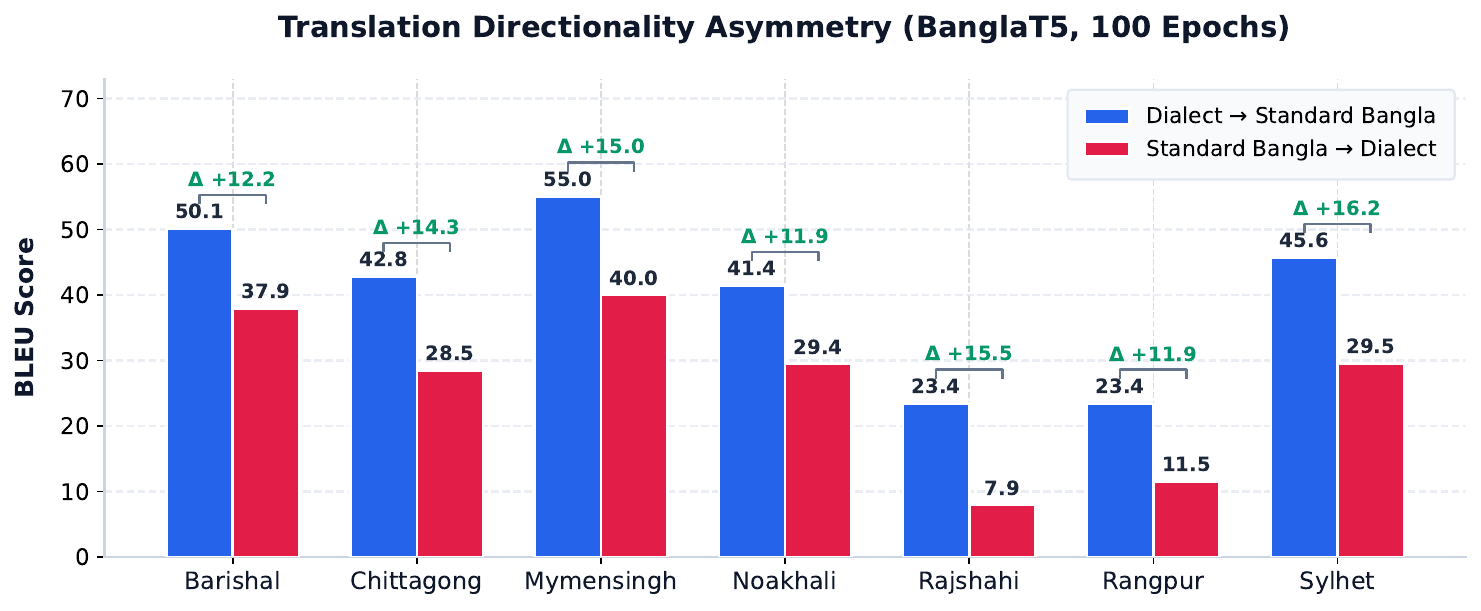}
  \caption{Translation directionality asymmetry across six major dialects. Dialect$\rightarrow$SCB (green) consistently outperforms SCB$\rightarrow$Dialect (blue), with the BLEU gap indicated by annotations.}
  \label{fig:directionality}
\end{figure}

\begin{enumerate}[leftmargin=*]
    \item \textbf{Impact of Linguistic Proximity and Data Volume:} Dialects that are linguistically closer to SCB yield substantially higher translation quality, although dataset volume also plays a secondary role (illustrated in Figure~\ref{fig:proximity_bleu}). To quantify this, the linguistic distance $d$ between a dialect $D$ and SCB $S$ is measured on a relative scale ($1$--$10$) based on Normalized Edit Distance (NED), defined as:
    \begin{equation}
        d(D, S) = 10 \times \frac{1}{|V|} \sum_{i=1}^{|V|} \frac{\text{Levenshtein}(w_{D}^{(i)}, w_{S}^{(i)})}{\max(|w_{D}^{(i)}|, |w_{S}^{(i)}|)}
    \end{equation}
    where $V$ is a common vocabulary set, $w_{D}^{(i)}$ and $w_{S}^{(i)}$ are corresponding lexical tokens in the dialect and SCB, and $\text{Levenshtein}(\cdot, \cdot)$ computes the minimum number of single-character edits required to transform one word into the other. 
    Mymensingh$\rightarrow$SCB achieves the highest pairwise BLEU of 55.0 ($d=2.5$), attributable to its conservative morphological divergence from standard forms---primarily suffix alterations rather than wholesale lexical replacement---despite having only a moderate dataset volume. Conversely, Chittagonian and Sylheti, which exhibit extensive lexical substitution and phonological restructuring ($d=9.0$ and $d=8.2$, respectively), produce lower BLEU scores (42.76 and 45.65 to SCB) even with substantially larger training sets. This finding corroborates dialectological classifications \cite{chatterji1926origin} and demonstrates that linguistic proximity often outweighs raw data volume in determining translation efficacy.
    \item \textbf{Directional Asymmetry:} Translating \textit{from} a regional dialect to SCB is consistently easier than the reverse direction (Figure~\ref{fig:directionality}). For example, Barisali$\rightarrow$SCB achieves 50.1 BLEU versus 37.9 for SCB$\rightarrow$Barisali---a gap of 12.2 points. This asymmetry reflects a fundamental decoder-side constraint: generating dialectal output requires producing highly localised morphological inflections and lexical items that are sparsely represented in the SCB-dominated pre-training corpus, whereas dialect-to-standard translation can exploit the model's strong prior over standard forms.
\end{enumerate}

\subsubsection{Qualitative Translation Examples}
Table~\ref{tab:qualitative} presents concrete translation examples demonstrating the model's capacity for syntactic reordering, lexical substitution, and morphological adaptation across diverse dialect pairs.

\begin{table}[htbp]
\centering
\caption{Qualitative translation examples generated by the final Poly-Dialectal BanglaT5 model across diverse regional dialect pairs, comparing generated output against reference ground truth.}
\label{tab:qualitative}
\vspace{0.2cm}
\includegraphics[width=1.0\textwidth]{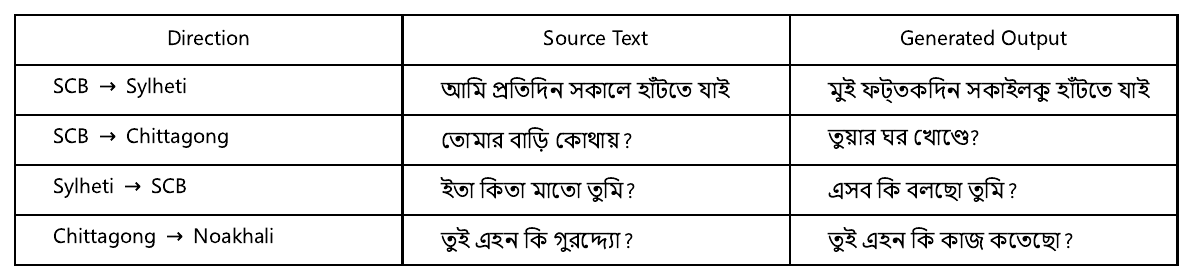}
\end{table}

The examples demonstrate that the model correctly performs lexical substitutions (e.g., \textit{ami}$\rightarrow$\textit{mui}, \textit{bari}$\rightarrow$\textit{ghor}), morphological inflection changes (e.g., \textit{protidin}$\rightarrow$\textit{fottokdin}), and direct dialect-to-dialect conversion without pivoting through SCB.

\subsubsection{Error Analysis}
\begin{figure}[htbp]
  \centering
  \includegraphics[width=0.75\textwidth]{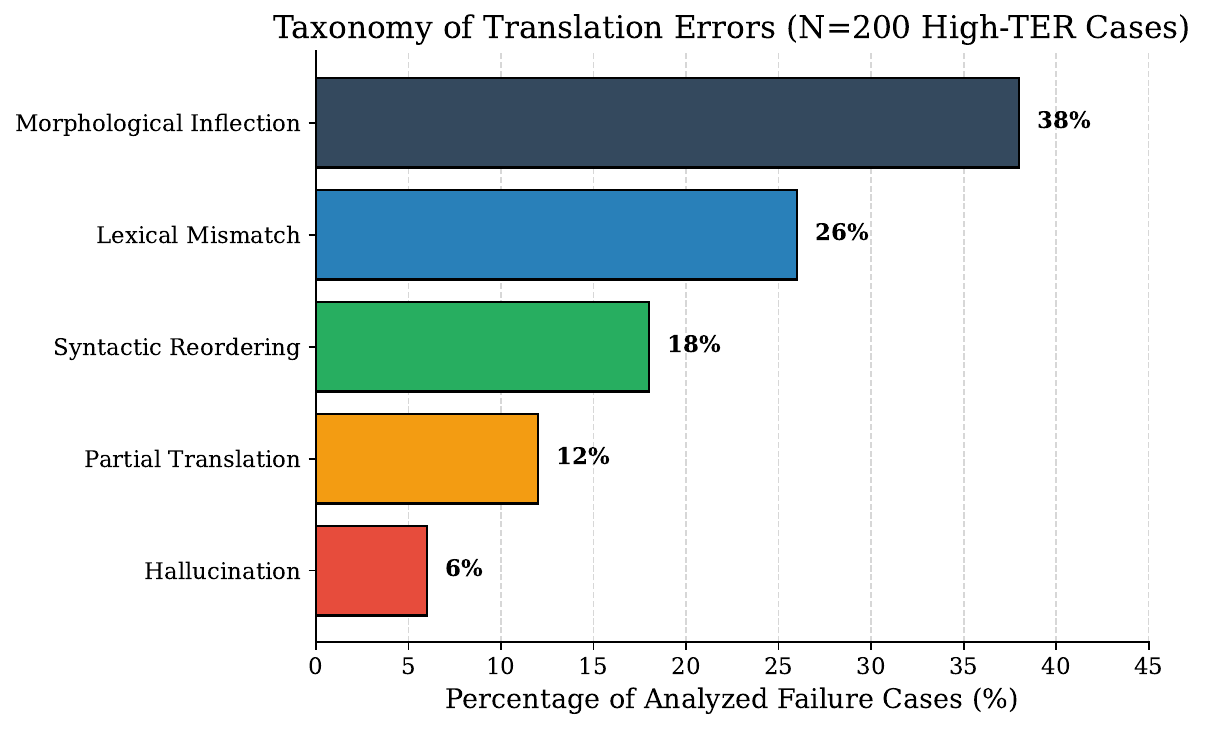}
  \caption{Distribution of translation errors across five linguistic failure categories, derived from manual analysis of 200 high-TER failure cases.}
  \label{fig:error_analysis}
\end{figure}

Despite high structural accuracy overall, the model exhibits systematic failure modes, particularly for low-resource dialect pairs. We manually analysed 200 failure cases (TER $> 80$), stratified across all 12 dialects to prevent over-representation of high-resource variants, and categorised the errors into five primary linguistic failure modes (Figure~\ref{fig:error_analysis}).

The dominant error categories are characterized as follows:
\begin{itemize}[leftmargin=*]
    \item \textbf{Morphological Inflection Errors (38\%):} The model predicts the correct root verb or noun but applies an incorrect regional suffix or inflectional ending. For example, generating \textit{korchi} (SCB progressive) instead of the correct Sylheti form \textit{koirddam}. This category is most prevalent in Chittagonian and Sylheti outputs, where verbal morphology diverges most sharply from SCB paradigms.
    \item \textbf{Lexical Mismatch (26\%):} The model substitutes a highly localized dialectal word with a semantically related but region-inappropriate SCB synonym. This occurs when the training data contains insufficient examples of region-specific vocabulary items.
    \item \textbf{Syntactic Reordering Failures (18\%):} Errors in constituent order, particularly in verb-final constructions characteristic of certain eastern dialects. The model occasionally retains SCB word order when the target dialect requires different sequencing.
    \item \textbf{Partial Translation (12\%):} The model produces an output that is partially translated, with some segments remaining in the source dialect. This is concentrated in low-resource pairs (e.g., Rajshahi$\rightarrow$Rangpur).
    \item \textbf{Hallucination (6\%):} The model generates tokens semantically unrelated to the source, typically occurring with extremely short source sentences or highly idiomatic expressions \cite{koehn-knowles-2017-six}.
\end{itemize}

\subsection{Dataset Scaling Analysis}
To quantify the relationship between training data volume and translation quality, we conducted a controlled scaling experiment using a fully parallel subset of four dialects (Sylheti, Chittagonian, Noakhali, Mymensingh) plus SCB, incrementally scaling from 500 to 4,499 sentence pairs under two DoRA configurations ($r=8, \alpha=16$ and $r=64, \alpha=128$).

\begin{figure}[h]
  \centering
  \includegraphics[width=0.90\textwidth]{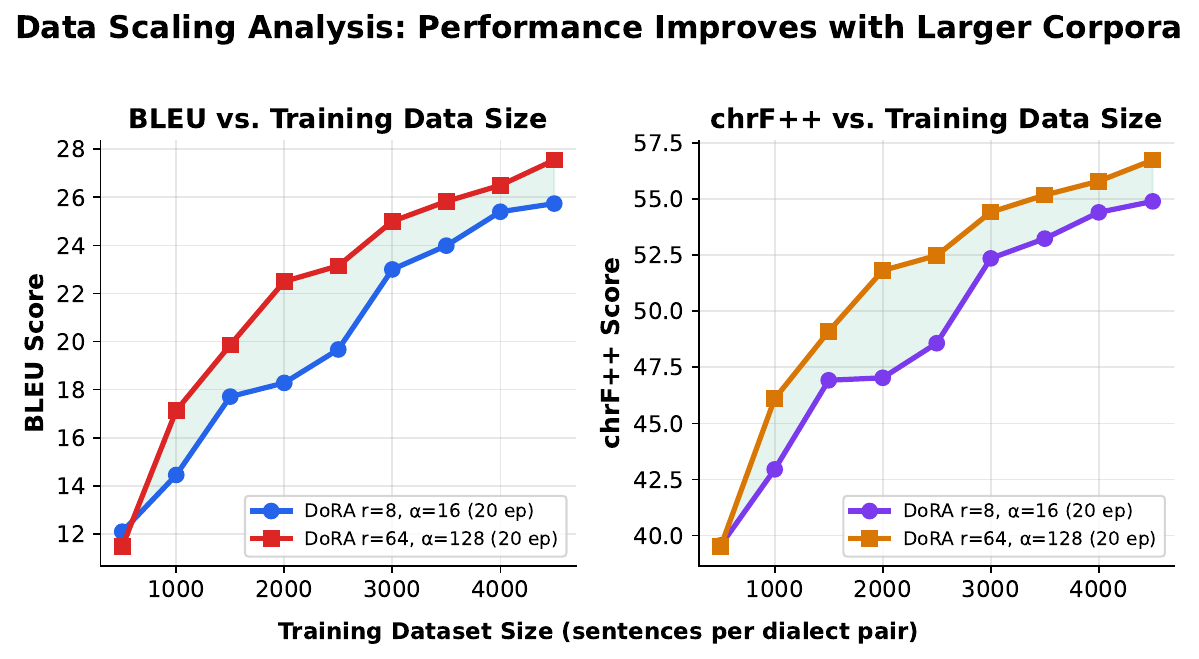}
  \caption{BLEU and chrF++ as a function of training data volume for two DoRA configurations. Consistent improvement is observed, with diminishing returns beyond approximately 3,000 samples.}
  \label{fig:scaling}
\end{figure}

As shown in Figure~\ref{fig:scaling}, BLEU improves by 113\% (from 12.09 to 25.74) as data scales from 500 to 4,499 pairs under the $r{=}8$ configuration, and by 140\% (from 11.47 to 27.55) under the $r{=}64$ configuration. Two key observations emerge. First, the steepest performance gains occur between 500 and 2,000 samples (BLEU approximately doubles), indicating that even modest data augmentation yields substantial returns for low-resource dialects. Second, both configurations exhibit diminishing returns beyond 3,000 samples, suggesting a practical saturation threshold for this corpus size \cite{koehn-knowles-2017-six}. The higher-capacity $r{=}64$ configuration consistently outperforms $r{=}8$ by 1--2 BLEU points, confirming that increased adapter capacity provides marginal but consistent benefits for morphologically complex dialectal translation.

\subsection{Comparison with Prior Studies}
To contextualize the proposed system, Table~\ref{tab:sota_comparison} benchmarks our results against recent state-of-the-art systems for Bangla dialect translation.

\begin{table}[htbp]
\centering
\caption{Comparison with existing state-of-the-art benchmarks. Direct comparison should be interpreted with caution, as different works use different test sets and dialect combinations.}
\label{tab:sota_comparison}
\resizebox{\textwidth}{!}{
\begin{tabular}{l l l l c c}
\toprule
\textbf{Work} & \textbf{Corpus Scale} & \textbf{Dialect Scope} & \textbf{Base Architecture} & \textbf{BLEU} & \textbf{chrF++} \\
\midrule
Vashantor \cite{faria2025vashantor} & 32,500 pairs & 5 Dialects & DialectBanglaT5 & 15.42 & 40.15 \\
BhasaBodh \cite{bhuiyan2025bhasabodh} & 1,960 pairs & 2 Dialects & mBART-50 & 17.80 & 43.50 \\
BanglaCHQ \cite{mahjabin2025banglachq} & 2,850 pairs & 2 Dialects & Gemini 2.5 Flash & 23.67 & 48.90 \\
\rowcolor{gray!15} \textbf{Proposed (Ours)} & \textbf{51,541 pairs} & \textbf{12 Dialects} & \textbf{BanglaT5 (100 ep)} & \textbf{29.26} & \textbf{57.26} \\
\bottomrule
\end{tabular}
}
\end{table}

\begin{figure}[htbp]
  \centering
  \includegraphics[width=0.75\textwidth]{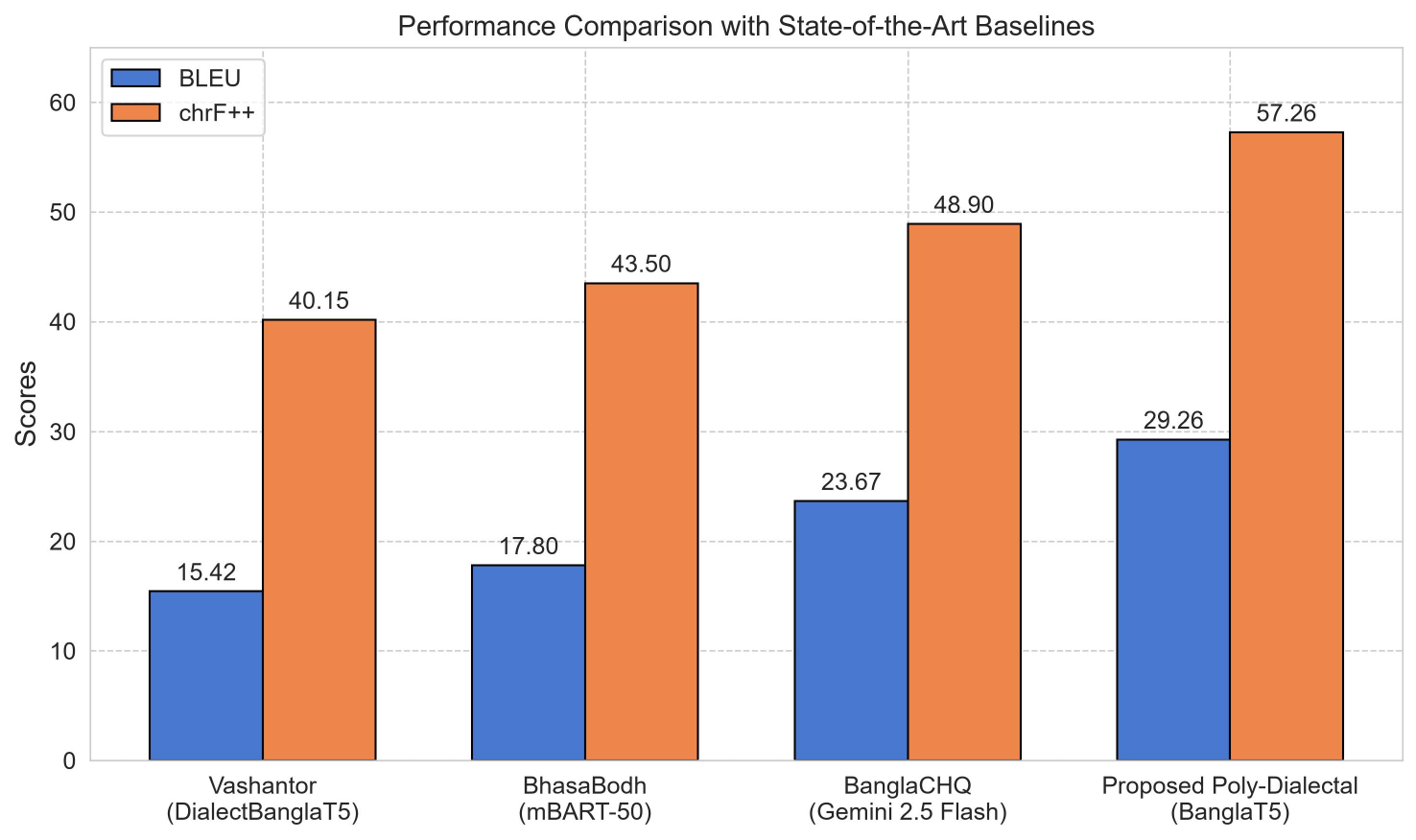}
  \caption{Visual comparison of BLEU and chrF++ scores across baseline systems and the proposed approach.}
  \label{fig:comparative_sota}
\end{figure}

The proposed system achieves a BLEU score 5.59 points higher than the best prior result (BanglaCHQ-Prantik, 23.67 BLEU using Gemini 2.5 Flash), representing a 23.6\% relative improvement (see Figure~\ref{fig:comparative_sota}). The chrF++ gain of 8.36 points (from 48.90 to 57.26) is particularly noteworthy, as chrF++ is better suited to morphologically rich languages where partial word matches carry semantic value. We note, however, that direct numerical comparison across studies should be interpreted cautiously, since each work employs different test sets, dialect combinations, and evaluation conditions. The primary advantage of the proposed approach lies not only in aggregate metric gains but also in the breadth of coverage (12 dialects vs.\ 2--5) and the unified multi-directional architecture that eliminates pivot-based translation. Moreover, whereas prior studies evaluate performance using only one or two metrics, our work assesses translation quality across four complementary metrics---including METEOR and TER---providing a more comprehensive characterisation of syntactic accuracy, semantic alignment, and practical post-editing effort.

\section{Deployment and Application Architecture}

To bridge the gap between academic research and real-world accessibility, the final poly-dialectal NMT system was deployed as an interactive web application, publicly available at \url{https://bangla-regional-translator.streamlit.app/}. The platform provides real-time translation across 12 dialects. The source code is available at 
\newpage
\url{https://github.com/secrakib/Defence_Translator_App}.

\subsection{Model Quantization and Serving}
\begin{figure}[h]
  \centering
  \includegraphics[width=0.5\textwidth]{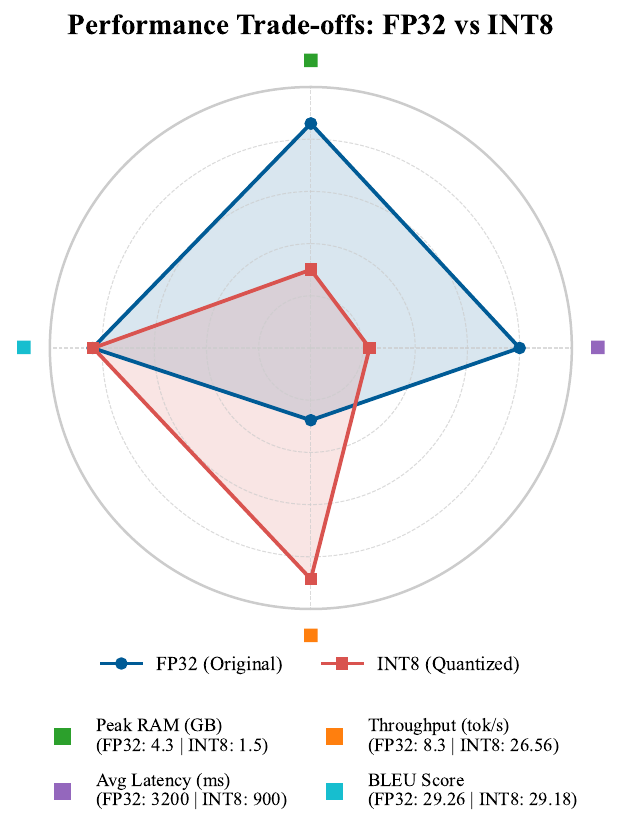}
  \caption{Multidimensional performance trade-off (Radar chart) demonstrating the advantages of INT8 quantization in reducing latency and memory footprint while preserving translation quality.}
  \label{fig:quantization}
\end{figure}

The primary challenge in deploying NMT systems to production environments is memory consumption. To address this, the fine-tuned BanglaT5 model with merged DoRA weights was converted to the \textbf{CTranslate2} \cite{klein-etal-2017-opennmt} inference engine---a C++ runtime optimised for Transformer architectures with hardware-accelerated vector operations (AVX2/AVX-512). Model weights were quantised from 32-bit floating-point to 8-bit integer (INT8) precision \cite{Jacob_2018_CVPR}:
\begin{equation}
W_{\text{INT8}} = \text{Quantize}(W_{\text{FP32}})
\end{equation}
This quantization reduced peak RAM consumption by over 65\% (operating under 1.5 GB) while accelerating CPU inference speed by approximately $3.2\times$ without perceptible degradation in translation quality (see Figure~\ref{fig:quantization}).

\subsection{Inference Pipeline and Long Document Processing}
Encoder--decoder Transformers enforce a fixed maximum input length ($L_{\max} = 128$ tokens in our configuration). To accommodate full-length articles, the backend implements a dynamic sentence segmentation pipeline that splits input text at boundary delimiters---the Bangla \textit{dairi} (``\textbar''), question marks, exclamation marks, and newlines. Each extracted sentence chunk is independently translated through the quantised engine using beam search decoding (beam size $B = 4$, length penalty $\alpha = 0.8$, repetition penalty $\beta = 1.2$) \cite{wu2016google}, and the translated chunks are concatenated sequentially to reconstruct the full document.

\subsection{Inference Latency Benchmarks}
Table~\ref{tab:latency} presents empirical latency and throughput measurements on standard cloud CPU instances.

\begin{table}[htbp]
\centering
\caption{End-to-end inference latency and throughput benchmarks for the deployed INT8-quantized model.}
\label{tab:latency}
\small
\begin{tabular}{l c c c c}
\toprule
\textbf{Input Type} & \textbf{Length} & \textbf{Tokens} & \textbf{Latency (ms)} & \textbf{Throughput (tok/s)} \\
\midrule
Short sentence & $\sim$10 words & 15 & $\sim$900 & 16.67 \\
Medium paragraph & $\sim$25 words & 38 & $\sim$1,450 & 26.20 \\
Long paragraph & $\sim$60 words & 85 & $\sim$3,200 & 26.56 \\
\bottomrule
\end{tabular}
\end{table}

\subsection{User Interface and Translation Demonstrations}
\begin{figure}[htbp]
  \centering
  \begin{subfigure}[b]{0.85\textwidth}
    \centering
    \includegraphics[width=\textwidth]{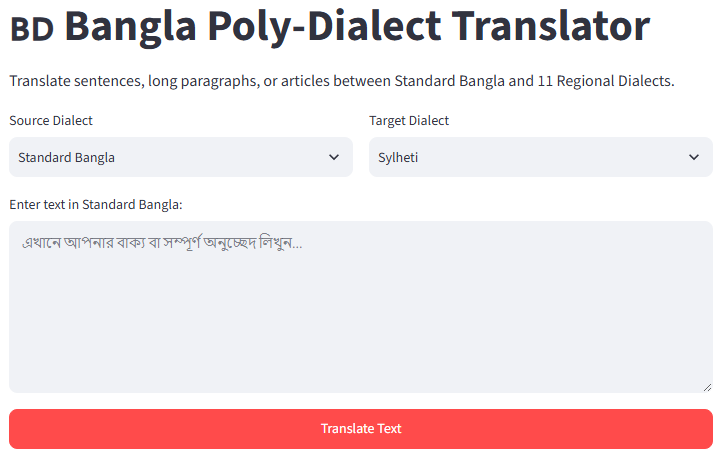}
    \caption{The deployed Streamlit web application interface showing source/target dialect selectors, text input area, and translation output panel.}
    \label{fig:deployment_ui}
  \end{subfigure}
  
  \vspace{0.4cm} 
  
  \begin{subfigure}[b]{0.48\textwidth}
    \centering
    \includegraphics[width=\textwidth]{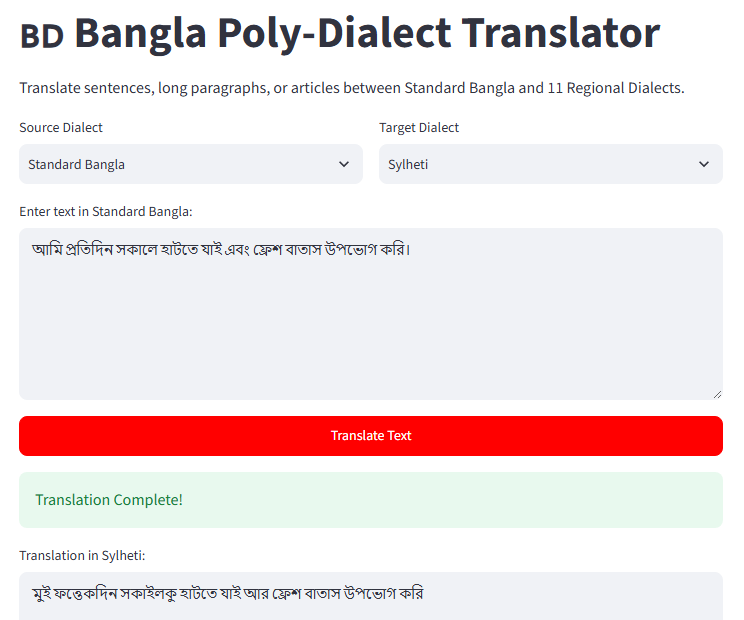}
    \caption{Standard Bangla to Sylheti translation demonstration, showing morphological and lexical adaptation.}
    \label{fig:deploy_sylheti}
  \end{subfigure}
  \hfill
  \begin{subfigure}[b]{0.48\textwidth}
    \centering
    \includegraphics[width=\textwidth]{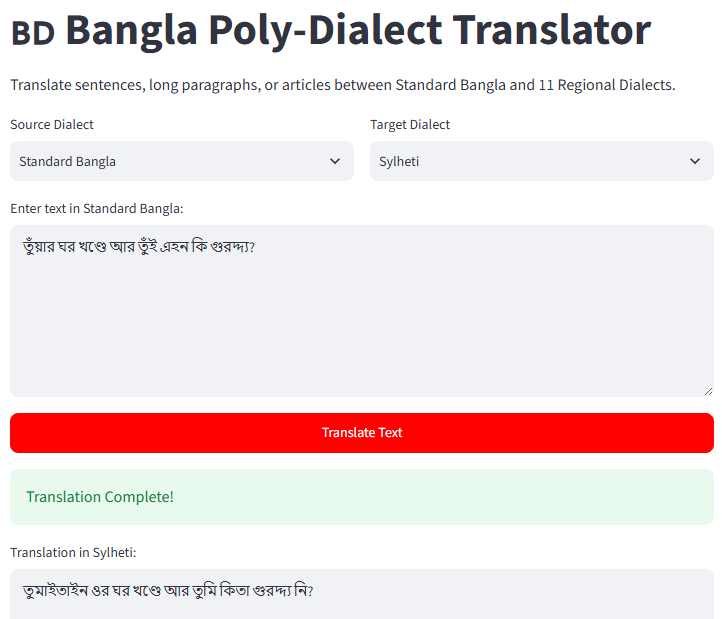}
    \caption{Direct dialect-to-dialect translation from Chittagong to Noakhali, bypassing the Standard Bangla pivot.}
    \label{fig:deploy_direct}
  \end{subfigure}
  
  \caption{Overview and translation inference demonstrations of the deployed multi-dialect translation system: (a) main interface, (b) standard-to-dialect translation, and (c) direct dialect-to-dialect translation.}
  \label{fig:deployment_overview}
\end{figure}

The application features a clean dual-column dialect selector layout with dynamic progress indicators. Figure~\ref{fig:deployment_ui} shows the main interface, while Figures~\ref{fig:deploy_sylheti} and~\ref{fig:deploy_direct} demonstrate representative translation outputs.

\section{Limitations and Future Work}

\subsection{Current Limitations}
Several limitations of the present work merit acknowledgment:

\begin{enumerate}[leftmargin=*]
    \item \textbf{Persistent Class Imbalance:} Despite integrating seven corpora and supplementing them with manual annotation, the dataset retains a 21:1 imbalance ratio between Standard Bangla (13,556 entries) and the least represented dialects (e.g., Narail and Tangail at 500 entries each). This imbalance constrains generalisation for underrepresented dialects, as evidenced by the substantially lower BLEU scores for Rajshahi (23.40) and Rangpur (23.39) relative to well-represented dialects such as Mymensingh (55.00) in the dialect-to-SCB direction.
    \item \textbf{Tokenisation Limitations:} Although BanglaT5's vocabulary outperforms its multilingual counterparts for Bangla text, its 32K SentencePiece vocabulary was trained exclusively on Standard Bangla and may still produce suboptimal segmentation for highly divergent dialectal morphemes. No dialect-specific tokeniser adaptation was performed in this study.
    \item \textbf{Text-Only Modality:} The system operates exclusively on orthographic text, disregarding the phonological richness and tonal variations that characterise spoken Bangla dialects. Many dialectal distinctions are primarily oral and are not fully captured by written representations.
    \item \textbf{Absence of Human Evaluation:} While automatic metrics provide reproducible baselines, they cannot fully capture semantic adequacy, fluency, and socio-linguistic authenticity as perceived by native speakers \cite{mathur-etal-2020-tangled}. The absence of formal human evaluation limits the interpretability of our quality assessments.
    \item \textbf{Evaluation Scope:} The automatic metrics used assess surface-level textual similarity and may not capture deeper pragmatic or cultural dimensions of translation quality. Additionally, because the test set is drawn from the same distribution as the training data, the results may not fully reflect the dialectal text variation encountered in informal digital communication.
\end{enumerate}

\subsection{Future Directions}
\begin{enumerate}[leftmargin=*]
    \item \textbf{Corpus Expansion via Community Annotation:} Launching community-driven annotation campaigns in collaboration with regional linguistic communities to systematically expand parallel datasets for underrepresented dialects. Crowdsourcing platforms with native-speaker verification could scale data collection to thousands of additional pairs per dialect.
    \item \textbf{Data Augmentation Strategies:} Investigating back-translation \cite{sennrich-etal-2016-improving}, paraphrasing, and dialect-conditioned synthetic data generation using large language models \cite{brown2020language, touvron2023llama} to augment low-resource dialect pairs without compromising data quality.
    \item \textbf{Multimodal Integration:} Extending the framework to incorporate automatic speech recognition (ASR) \cite{radford2023robust} and text-to-speech (TTS) modules, enabling end-to-end spoken dialect translation. This is particularly relevant for illiterate populations who communicate primarily in spoken dialect.
    \item \textbf{Formal Human Evaluation:} Executing a large-scale human evaluation study employing the Multidimensional Quality Metrics (MQM) framework \cite{burchardt-2013-multidimensional} or Direct Assessment protocols \cite{graham2017machine} with native linguists to assess translation fluency, adequacy, and dialectal authenticity.
    \item \textbf{Cross-Lingual Transfer:} Exploring transfer learning from related Indo-Aryan languages (e.g., Assamese, Hindi regional dialects) that share typological features with Bangla dialects, potentially improving performance for the most divergent variants.
\end{enumerate}

\section{Conclusion}

This paper presented a comprehensive framework for poly-dialectal neural machine translation in Bangla, directly addressing the technological gap between the language's rich dialectal diversity and the monolithic character of existing NLP systems. The principal contributions and findings are summarised below.

First, we constructed the \textbf{largest multi-dialectal parallel corpus} for Bangla machine translation to date, comprising 51,541 non-null parallel sentence pairs across 12 dialects. The corpus integrates seven existing datasets and introduces 2,510 expert-verified sentence pairs for five previously unaddressed dialects (Rangpur, Tangail, Kishoreganj, Narail, Narsingdi), providing up to 2.76$\times$ more data per dialect than prior efforts.

Second, systematic benchmarking demonstrated that language-specific pre-training provides a substantially stronger foundation for intra-language dialectal translation than massively multilingual models. The 247M-parameter BanglaT5 outperformed NLLB-200 (615M) by 51.8\% in BLEU and mBART-50 (611M) by 145.7\%. Extended 100-epoch DoRA fine-tuning yielded state-of-the-art scores of 29.26 BLEU and 57.26 chrF++, surpassing the best prior result \cite{mahjabin2025banglachq} by 5.59 BLEU points. Cross-dialectal analysis revealed that morphological proximity to Standard Bangla is the primary determinant of translation quality---Mymensingh achieved 55.0 BLEU to SCB whereas Chittagonian, with substantially larger training data, reached 42.76. A controlled dataset scaling study further established that translation quality improves by up to 140\% as parallel data scales from 500 to 4,499 pairs, with diminishing returns near the 3,000-sample mark.

Third, the system was deployed as a publicly accessible, INT8-quantised web application supporting real-time, multi-directional translation across 12 dialects---including direct dialect-to-dialect paths that eliminate the cascading errors of pivot-based methods. This deployment constitutes a concrete step toward digital inclusion and linguistic accessibility for millions of Bangla dialect speakers who currently lack computational language support.

\section*{Data Availability Statement}
The multi-dialect parallel corpus constructed, annotated, and benchmarked in this paper is publicly available on Mendeley Data at \url{https://data.mendeley.com/datasets/v9cf66fk2t}.

\section*{Ethics Statement}
All human annotators were compensated fairly according to local standards, and the corpus contains no personally identifiable information.

\bibliographystyle{plain}
\bibliography{biblography}

\end{document}